\pdfoutput=1
\PassOptionsToPackage{prologue,dvipsnames}{xcolor}

\documentclass[11pt]{article}

\usepackage[final]{acl}
\usepackage[dvipsnames]{xcolor}
\usepackage{times}
\usepackage{latexsym}
\usepackage{makecell}
\usepackage{CJKutf8}
\usepackage[T1]{fontenc}

\usepackage[utf8]{inputenc}
\usepackage{booktabs}
\usepackage{enumitem}
\usepackage{multirow}
\usepackage{colortbl}
\usepackage{tikz}
\usepackage{amssymb}
\usepackage{microtype}
\usepackage{soul}
\usepackage{subcaption}
\usepackage{inconsolata}

\usepackage{graphicx}
\usepackage{xspace}
\usepackage{float}
\usepackage{amsmath}
\usepackage{comment}
\usepackage{arydshln}
\usepackage{float,lscape}
\usepackage{afterpage}
\usepackage[colorinlistoftodos,prependcaption,textsize=tiny]{todonotes}
\usepackage{scalerel}

\newcommand{\se}[1]{\textcolor{black}{#1}}
\newcommand{\zr}[1]{\textcolor{black}{#1}}

\title{\transpro: An LLM-based \textsc{Lit}erary \textsc{trans}lation Evaluation Metric with 
\textsc{Pro}fessional \textsc{Q}uestion \textsc{A}nswering}

\author{Ran Zhang \\
Natural Language Learning Group (NLLG) \\
School of Business Informatics and Mathematics \\
University of Mannheim \\
\texttt{ran.zhang@uni-mannheim.de} 
 \And  ... \And
         Author n \\ Address line \\ ... \\ Address line}
\author{
  \textbf{Ran Zhang\textsuperscript{1,5}},
  \textbf{Wei Zhao\textsuperscript{2}},
\textbf{Lieve Macken\textsuperscript{3}},
 \textbf{Steffen Eger\textsuperscript{4,5}},
\\
  \textsuperscript{1}University of Mannheim, School of Business Informatics and Mathematics \\
  \textsuperscript{2}University of Aberdeen, Department of Computing Science\\
  \textsuperscript{3}University of Gent, Department of Translation, Interpreting and Communication \\
  \textsuperscript{4}University of Technology Nuremberg (UTN), Department Engineering
  \\ \textsuperscript{5}Natural Language Learning and Generation (NLLG) Lab
\\
  \small{
    \textsuperscript{1} \href{mailto:ran.zhang@uni-mannheim.de}{ran.zhang@uni-mannheim.de} \ \ 
    \textsuperscript{2}  \href{mailto:wei.zhao@abdn.ac.uk}{wei.zhao@abdn.ac.uk} \ \ 
    \textsuperscript{3}  \href{mailto:lieve.macken@ugent.be}{lieve.macken@ugent.be} \ \ 
    \textsuperscript{4}  \href{mailto:steffen.eger@utn.de}{steffen.eger@utn.de}
  }
}
\newcommand{\liteval}{\textsc{LitEval-Corpus}\xspace}
\newcommand{\qmono}{\textsc{1/4-FrEn}\xspace}
\newcommand{\hmono}{\textsc{1/2-FrEn}\xspace}
\newcommand{\qmulti}{\textsc{1/4-multi}\xspace}
\newcommand{\hmulti}{\textsc{1/2-multi}\xspace}
\newcommand{\vawei}{$\text{Vanilla}_{\text{w}}$\xspace}
\newcommand{\va}{$\text{Vanilla}$\xspace}
\newcommand{\pstepwei}{$\text{PromptStep}_{\text{w}}$\xspace}
\newcommand{\pstep}{$\text{PromptStep}$\xspace}
\newcommand{\qstepwei}{$\text{QuestionStep}_{\text{w}}$\xspace}
\newcommand{\qstep}{$\text{QuestionStep}$\xspace}
\newcommand{\cometxl}{\textsc{XCOMET-xl}\xspace}
\newcommand{\cometkiwi}{\textsc{COMET-kiwi}\xspace}
\newcommand{\acceq}{\textsc{Acc-eq}\xspace}
\newcommand{\kendall}{Kendall's\ $\tau$\xspace}
\newcommand{\deepl}{DeepL\xspace}
\newcommand{\gtran}{Google\ Translate\xspace}
\newcommand{\gtr}{\textsc{GTR}\xspace}
\newcommand{\gptfo}{GPT-4o\xspace}
\newcommand{\gpttf}{GPT-3.5\xspace}
\newcommand{\llama}{LLaMA}

\newcommand{\tower}{TowerInstruct\xspace}

\newcommand{\nllb}{NLLB\xspace}
\newcommand{\qwen}{Qwen\xspace}
\newcommand{\gemba}{\textsc{GEMBA-MQM}\xspace}
\newcommand{\deen}{De-En\xspace}
\newcommand{\enDe}{En-De\xspace}
\newcommand{\enzh}{En-Zh\xspace}
\newcommand{\dezh}{De-Zh\xspace}
\newcommand{\cometxxl}{\textsc{XCOMET-xxl}\xspace}
\newcommand{\iyyer}{\textsc{LiteraryTran}\xspace}
\newcommand{\sent}{\textsc{sent}\xspace}
\newcommand{\sentp}{\textsc{para\_sent}\xspace}
\newcommand{\para}{\textsc{para}\xspace}
\newcommand{\paran}{\textsc{PAR3-annotated}\xspace}
\newcommand{\parun}{\textsc{PAR3-unannotated}\xspace}

\newcommand{\transpr}{\textsc{LiTransProQA}\xspace}
\newcommand{\transpro}{\texorpdfstring{\transpr}{TransPro}}

\begin{document}
\maketitle
\begin{abstract}
The impact of Large Language Models (LLMs) has extended into literary domains. However, existing evaluation metrics \se{for literature} prioritize mechanical accuracy over artistic expression and tend to overrate machine translation as being superior to human translation from experienced professionals. In the long run, this bias could result in an irreversible decline in translation quality and cultural authenticity. In response to the urgent need for a specialized literary evaluation metric, we introduce \transpro, a novel, reference-free, LLM-based question-answering framework designed for literary translation evaluation. \transpro integrates humans in the loop to incorporate insights from professional literary translators and researchers, focusing on critical elements in literary quality assessment such as literary devices, cultural understanding, and authorial voice. Our extensive evaluation shows that while literary-finetuned \cometxl yields marginal gains, \transpro substantially outperforms current metrics, achieving up to 0.07 gain in correlation and surpassing the best state-of-the-art metrics by over 15 points in adequacy assessments. Incorporating professional translator insights as weights further improves performance, highlighting the value of translator inputs. Notably, \zr{\transpro reaches an adequacy performance comparable to trained linguistic student evaluators, though it still falls behind experienced professional translators.} \transpro shows broad applicability to open-source models like \llama3.3-70b and Qwen2.5-32b, indicating its potential as an accessible and training-free tool for evaluating literary translations that require local processing due to copyright or ethical considerations.
\end{abstract}
\section{Introduction}\label{sec:intro}
Large Language Models (LLMs) have shown remarkable capabilities in linguistic tasks, emerging as potentially transformative tools across many domains \cite{grattafiori2024llama, yang2024qwen2, achiam2023gpt, Eger2025TransformingSW}. However, their suitability for more nuanced creative areas---such as
literary translation or poetry generation---remains uncertain \cite{macken_2024_LLMmachine, chakrabarty2024art, al2021translation, chen2024evaluating, belouadi_2023_bygpt5, zhang2024llm}.
\begin{figure*}
\centering
\includegraphics[width=0.98\linewidth]{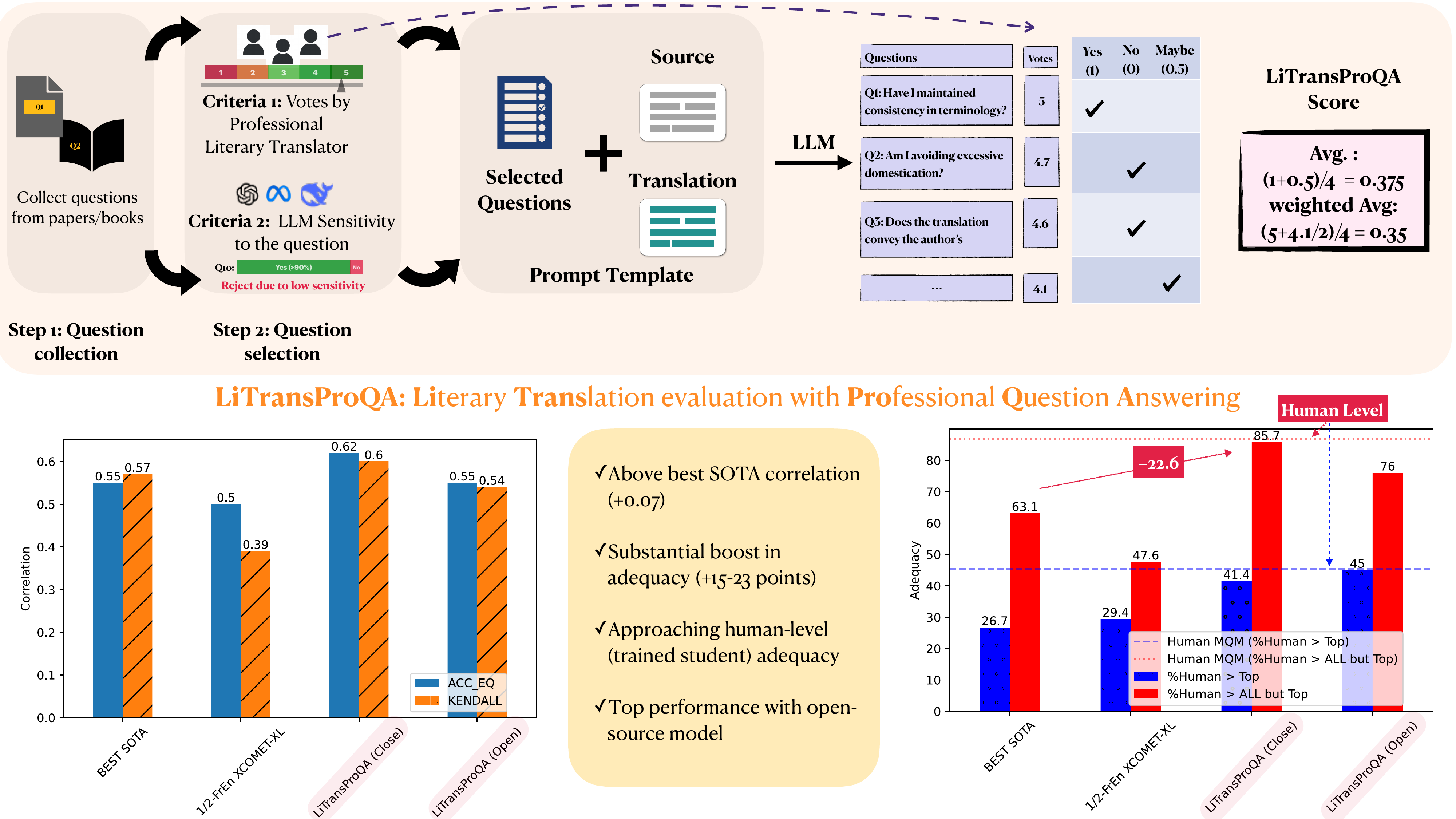}
\caption{Overview of \transpro and its performance compared to finetuned and LLM-based SOTA in correlation and adequacy (the ability to rate high-quality human translation better than MT). Human MQM (dashed lines) represents the adequacy level of trained students using MQM. \zr{Human level refers to trained student evaluators, who still lag behind experienced professional translators by a large margin according to Zhang et al. (2024).}}
\label{fig:fig1}
\end{figure*}
Literary translation requires not just lexical and syntactic precision, but also a deep understanding of cultural context, aesthetic style, and interpretive nuance \cite{wang_2023_findings, karpinska2023large, matusov_2019_challenges, pang2024salute}. To critically assess the suitability of LLMs for such complex creative work, it is essential to establish robust and systematic \emph{evaluation methods} that can truly capture the essence of literary translation.

While human evaluation appears ideal for assessing literary translation qualities, it becomes economically impractical at scale, given the vast corpus of world literature and LLMs' unprecedented generation capabilities. Moreover, proper evaluation requires input from trained literary professionals, making human assessment prohibitively expensive \cite{zhang2024good, yan2024gpt}.

The inherent nature of literary translation compounds this evaluation challenge. While technical texts often have clear ``correct'' translations, literary works demand creative reinterpretation across linguistic and cultural boundaries. This poses fundamental problems for reference-based evaluation methods, as generating reference translations for literary texts is not only resource-intensive but also conceptually problematic, since multiple valid interpretations can exist simultaneously.

Existing automatic evaluation approaches fall short fundamentally: previous metrics like BLEU \cite{papineni2002bleu} and METEOR \cite{banerjee2005meteor}, and later embedding-based approaches such as BERTScore \cite{Zhang*2020BERTScore} and BLEURT \cite{sellam-etal-2020-bleurt} primarily measure semantic equivalence and linguistic accuracy, overlooking core literary attributes, such as tone, cultural specificity, and figurative language.

Even recent state-of-the-art (SOTA) automatic evaluation approaches, including \cometxl \& \cometxxl \cite{xcomet_tacl_a_00683}, LLM-based metrics like \gemba \cite{kocmi2023gemba}, and hybrid LLMs like Prometheus \cite{kim2024prometheus}, show limitations in literary contexts, despite their design to handle reference-free evaluation and potentially cover surface-level stylistic elements. \citet{zhang2024good} demonstrate that these metrics may consistently prefer machine-generated literary translations over translations written by experienced professional translators, which severely misaligns with judgments from human experts. The lack of professional translation expertise in current metrics, particularly regarding literary translation nuances and quality assessment standards, contributes to their limited ability to conduct meaningful evaluations.

This paper addresses these limitations by introducing \transpro, as shown in Figure \ref{fig:fig1}, a novel LLM-based \textsc{lit}erary \textsc{Trans}lation evaluation metric focusing on \textsc{Pro}fessional \textsc{q}uestion \textsc{a}nswering. \transpro integrates humans in the loop to reflect professional translators' quality control and assessment processes.\footnote{The code and datasets are available under: \url{https://github.com/NL2G/LiTransProQA}.} Unlike existing QA-based MT metrics \cite{krubinski-etal-2021-just, fernandes2025llms}, \transpro focuses on core elements in literary translation proposed and verified by researchers and professional literary translators for better alignment with human experts. Our work presents a detailed question development and selection process for an optimized set of questions, as our analysis indicates that LLMs are not yet fully trustworthy for automatic question generation and QA evaluation in the literary domain.

In addition to \transpro, we finetune \cometxl, one of the dominant metrics for standard MT, with literary domain tasks for comparison. We evaluate metric performance on carefully selected human-annotated datasets where we find that (1) while finetuning \cometxl yields only marginal gains, \transpro delivers substantial performance gains even compared to the best SOTA in both correlation with human judgments and adequacy in rating published human translations as better than MT outputs, with gains of nearly 0.07 in correlation measured by \acceq and \kendall and over 15 points in adequacy; (2) incorporating translator votes improves \transpro's performance, showing the value of professional input in evaluation; (3) \zr{\transpro demonstrates substantial progress in adequacy, approaching the level of trained linguistic student annotators though still lagging behind experienced professionals, which highlights its potential and room for improvements}; and 
(4) \transpro shows robust performance using open-source models, demonstrating its value as an accessible, training-free metric for evaluating literary texts---particularly those requiring local processing for copyright or ethical reasons.
\section{Background \& related work}\label{sec:related}
\paragraph{Dataset for literary translation}
Several large-scale \textbf{parallel corpora} exist in the literary domain. \textit{BWB} \cite{jiang2022blonde} and \textit{GuoFeng} \cite{xu2022guofeng} contain recent Chinese-English web novels, but their unspecified human reference quality and potential use of post-edited MT \cite{kolb2023bit} make them unsuitable for our study. \textit{PAR3} \cite{thai2022exploring}, another large-scale paragraph-level multilingual-to-English parallel corpus, includes 
published human translations. Although PAR3 lacks detailed metadata, its human translations can be manually verified \cite{zhang2024good}.

Existing \textbf{evaluation corpora with human judgments}, such as the WMT shared tasks, mainly focus on technical or news domains \cite{specia-etal-2020-findings,specia-etal-2021-findings, zerva-etal-2022-findings,blain-etal-2023-findings}---content that differs substantially from literary texts \cite{voigt_jurafsky_towards, matusov_2019_challenges,macken_2024_LLMmachine, van2023riddle}. The recent WMT24 shared task, which consists of evaluation corpora with human judgments, includes literary samples across 7 language pairs from English. However, their human annotation results show MT systems outperform human translations in 4 out of 7 language pairs, likely due to suboptimal human references from less experienced or non-literary translators \cite{zerva-etal-2024-findings}. Three additional datasets, i.e., \iyyer \cite{karpinska2023large}, \liteval \cite{zhang2024good}, and \paran, contain published human references with identifiable translators. While WMT24 and \liteval contain numerical scores as quality annotations, \iyyer and \paran contain pairwise preference rankings via direct comparison.

\paragraph{Automatic metrics for translation evaluation}
MT evaluation has evolved substantially from lexical overlap metrics like BLEU \cite{papineni2002bleu} to more sophisticated approaches using embeddings from pretrained models such as BERTScore \cite{Zhang*2020BERTScore},
Natural language inference-based MENLI \cite{chen2023menli}, or discourse-based DiscoScore \cite{zhao2023discoscore}. The COMET series \cite{xcomet_tacl_a_00683} mark a breakthrough in finetuning-based evaluation frameworks by effectively modeling the relationship between source and candidate translation (and reference) to better align with human judgments on translation quality. Moreover, the rise of LLMs has enabled powerful prompting metrics such as \gemba and finetuned LLM Prometheus \cite{kim2024prometheus}, pushing the boundaries of general-domain MT evaluation.

QA-based metrics present another promising direction, showing potential in various NLP tasks, including summarization \cite{fables-2024-kim-et-al, fabbri-etal-2022-qafacteval} and translation evaluation. For MT, SimQA \cite{han-etal-2022-simqa}, MTEQA \cite{krubinski-etal-2021-just}, AskQE \cite{ki2025askqe}, and TREQA \cite{fernandes2025llms} all follow a similar process: generating questions, obtaining answers, and evaluating answers. While SimQA and MTEQA (both reference-based) focus on keyphrase-driven question generation, AskQE and TREQA emphasize direct question generation from candidates and source context, both relying on comparisons between LLM-generated ground-truth and candidate answers. QA metrics specifically designed for literary text remain unexplored. Though TREQA includes evaluation on literary translation, it primarily focuses on measuring meaning and information accuracy, missing the nuanced literary elements discussed above. In our work, we finetune \cometxl with literary tasks and propose \transpro, a QA-based metric that largely differs from previous attempts: (1) rather than relying on automatic question generation, we collaborate with professional literary translators and researchers to develop a specialized set of questions. This approach ensures better alignment with the expertise and practices of domain experts and practitioners; (2) instead of using open-ended generated answers, our question design simplifies responses to \textit{yes, no, or maybe} to avoid overwhelming and confusing LLMs \cite{kamoi-etal-2023-shortcomings}; and (3) our answer evaluation does not rely on a generated ground-truth whose reliability remains untested \cite{huang2025survey}.
\section{Dataset}
We begin by introducing the datasets for metric evaluation (\liteval, \iyyer, and \paran) and for metric finetuning (WMT24 and \parun). Additional details and statistical summaries are provided in Section \ref{apdx:dataset} and Table \ref{tab:dataset} (appendix).

\subsection{Evaluation dataset}
We use 3 human-annotated datasets with verified published human references as evaluation sets.\footnote{\zr{For more details, see Section \ref{apdx:dataset} (appendix).}}

\paragraph{\liteval}
\liteval is a benchmark dataset for examining metric performance on literary translation evaluation. It contains paragraph-level parallel data with verified high-quality human translations across four language pairs: German–English (\deen), English–German (\enDe), German–Chinese (\dezh), and English–Chinese (\enzh), comprising over 2k paragraphs. The corpus includes outputs from 9 MT systems. Both human and MT translations are annotated with SOTA human evaluation scheme, Multidimensional Quality Metrics (MQM) \cite{freitag_2021_experts, lommel2014using}, allowing us to examine how metrics correlate with human judgments.\footnote{\zr{Our analysis uses the complete \liteval annotation dataset, created by four student evaluators with linguistics or translation study backgrounds. These evaluators are all native speakers of the target language. While \liteval also contains annotations from professional literary translators, this professional dataset is limited in size and covers only 3 language pairs. Therefore, we primarily use the student annotations for our correlation analysis and as our main human performance benchmark.}} This correlation is measured using \kendall and its variant \acceq \cite{deutsch-etal-2023-ties} with the official packages from WMT. The dataset includes human translations, enabling examination of metric \emph{adequacy}---how well metrics rank human translations above MT following \citet{zhang2024good}. We compare human translations against three cases: (1) the top 4 MT systems (\gptfo, \deepl, \gtran, \qwen2) as identified by \citet{zhang2024good} (top-level adequacy, the most challenging case); (2) all 9 MT systems (overall adequacy); and (3) all MT systems but top performers (low-level adequacy, the simplest case).

\paragraph{\iyyer}
\iyyer is a multilingual evaluation dataset containing source paragraphs from contemporary literature in English (En), German (De), French (Fr), Russian (Ru), Czech (Cs), and Japanese (Ja), with target translations in English, Japanese, and Polish (Pl). The dataset contains outputs from two MT systems: \gpttf under three prompting methods and \gtran. The dataset contains 540 direct pairwise preference annotations (1,080 source-target pairs). We compute the \acceq and \kendall of the metric against pairwise human preference judgments. The dataset also includes 180 human references for adequacy examination.

\paragraph{\paran}
\paran covers 3 language pairs (Fr-, Ru-, and De-En) from the PAR3 parallel corpus with direct preference annotations. The dataset includes translations from 2 MT systems (\gtran and \gpttf). Notably, \paran uses monolingual experts (writers and editors) for evaluation rather than translation professionals or linguists. From the dataset's 450 comparison cases (900 source-target pairs), we use 372 (744 pairs), excluding cases where evaluators rate \gtran or \gpttf outputs over human translations.

\subsection{Finetuning dataset}
We use 2 datasets to finetune \cometxl, each corresponding to a different literary task.
\paragraph{\parun: pairwise ranking task}
We utilize \parun for a literary ranking task. The corpus comprises classic literary paragraphs with human translations and \gtran outputs, covering translation pairs spanning both close (e.g., Fr-En) and distant language pairs (e.g., Zh-En). Its extensive size and diverse collection of human-translated literary texts make it ideal for model finetuning and domain adaptation. To expand our comparison between human and machine translations, we augment the dataset with outputs from smaller SOTA LLMs (GPT-4o-mini, TowerInstruct-13b/7b, Qwen2.5-7b, and \llama3-8b). We select these models for their cost efficiency and clear quality distinction from human translations, creating an effective ranking task for finetuning \cometxl. These machine-generated translations are paired with their corresponding human translations, as shown in Table \ref{tab:par3pair} (appendix).

\paragraph{WMT24: regression task}
WMT24 consists of evaluation corpora with human judgment using error span annotation (ESA) proposed recently by \citet{kocmi-etal-2024-error}. As discussed in Section \ref{sec:related}, it contains literary samples across 7 language pairs from English with 8-13 MT systems. We exclude samples with fewer than 10 tokens, resulting in 4,500 source-target pairs. We use this human-annotated corpus as a regression finetuning task.
\section{Experiment Design}
\begin{table}\centering
\scriptsize
\begin{tabular}{cccc}\toprule
\multicolumn{3}{c}{\textbf{1. Finetuning method}} \\\cmidrule{1-3}
\multirow{2}{*}{\textbf{finetuning task}} &\multicolumn{2}{c}{\textbf{finetuned layers}} \\\cmidrule{2-3}
&\textbf{\makecell{quarter \\(20.4\%, layer 28-36)}} &\textbf{\makecell{half \\ (40.7\%, layer 19-36)}} \\\midrule
\textbf{Reg.} & \textsc{1/4}-WMT24 & -\\
\textbf{Bi-ranking} &\qmono &\hmono \\
\textbf{Bi-ranking + Reg.} &\qmono-WMT24 & - \\
\textbf{Multi-ranking} &\qmulti &\hmulti \\
\textbf{Multi-ranking + Reg.} &\qmulti-WMT24 & - \\
\midrule
\multicolumn{3}{c}{\textbf{2. Prompting method}} \\
\midrule
\multirow{2}{*}{\textbf{prompt design}} &\multicolumn{2}{c}{\textbf{weighted by translator votes}} \\\cmidrule{2-3}
&\textbf{Yes} &\textbf{No} \\
\midrule
\textbf{Vanilla} & \vawei& \va \\
\textbf{\makecell{+ prompt instruction}} &\pstepwei & \pstep \\
\textbf{\makecell{+ question instruction}} & \qstepwei & \qstep \\
\bottomrule
\end{tabular}
\caption{Experimental setup for finetuning and prompting methods. Reg.\ stands for regression task using WMT24 dataset. Bi- and Multi-ranking stand for bi- and multilingual ranking tasks using \parun Fr-En and XX-En datasets, respectively.}\label{tab:exp_setup}
\end{table}
We introduce our metric development methods in the following section: (1) finetuning \cometxl with literary tasks and (2) developing an LLM-based QA metric reflecting the quality assessment process employed by professional literary translators. Table \ref{tab:exp_setup} summarizes the methodologies.

\subsection{Finetuning \cometxl}
\cometxl, built on the pretrained RoBERTa-XL model of 3b parameters, is finetuned on parallel translations with human judgments and error labels to predict quality scores given source-translation(-reference). However, since \cometxl is mainly trained on non-literary texts like news, it lacks domain knowledge in literary translation evaluation \cite{zhang2024good}. To address this, we finetune it on literary datasets using (1) a ranking task and (2) a regression task.

\paragraph{Finetuning setup}
For the \textbf{ranking task}, we use triplet training loss to finetune the \cometxl encoder, positioning human translations closer to their source texts than machine translations in the embedding space. For the \textbf{regression task}, we employ mean squared error loss to align quality assessment predictions with human-annotated scores.
We selectively finetune specific model layers to adapt \cometxl for literary domain while preserving its capabilities. For the ranking task, we test 2 layer-wise configurations: finetuning the top quarter layers (20.4\%, 28–36) and top half (40.7\%, 19–36). In parallel, we test 2 dataset configurations with 50k-paragraph translations: (1) a bilingual Fr-En \parun (the most common source language pair in PAR3) and (2) a multilingual \parun with various language pairs (XX-En). These configurations create four variants, shown in Table \ref{tab:exp_setup}: quarter- \& half-layer Fr-En (\qmono \& \hmono) and quarter- \& half-layer multilingual (\qmulti \& \hmulti). For regression, we finetune only the top quarter layers of \cometxl or of the finetuned \qmono \& \qmulti using WMT24. This setup helps evaluate how dataset diversity, layer depth, and different tasks affect performance. The finetuning parameters are reported in Table \ref{tab:param} (appendix).

\subsection{QA-based \transpro}
In addition to finetuning \cometxl, we introduce an LLM-based QA metric that reflects the professional translator's quality control and assessment process. \transpro consists of two key components: a prompting template and a question list paired with translator votes.

\paragraph{Template}
The \va template, shown in Table \ref{tab:template},  follows a simple structure: We first instruct LLM to be a professional literary translator. Next, we present a source-translation pair. Finally, we provide a list of evaluation questions that mirror the quality checks professional translators perform. The LLM answers each question with \textit{Yes, No} or \textit{Maybe}, which we map to scores of \textit{1, 0, or 0.5} respectively. Each translation receives a list of scores corresponding to the list of questions. The overall translation score is calculated as either an unweighted or translator-vote-weighted mean (denoted with $_w$). We also include template variations by introducing stepwise instructions to \va version or to individual questions to test whether more specific instructions could enhance performance.
\begin{table}\centering
\scriptsize
\begin{tabular}{m{2.6cm}m{4.6cm}}\toprule
\textbf{\va }& \textbf{\pstep} \\
\midrule
You are a professional literary translator with extensive experience. Now you are translating a work of great aesthetic value and cultural significance. You need to check if the translation covers all translation aspects by answering YES, NO or MAYBE to the following questions. Please be honest with your assessment and consider all aspects of translation quality. & 
\makecell[{{p{4.6cm}}}]{You are a professional literary translator with extensive experience. Now you're translating a work of great aesthetic value and cultural significance. You need to check if the translation covers all translation aspects by answering YES, NO or MAYBE to the following questions. 
\\
\\For each of the questions, 
\\1. Please first identify key translation components related to the question such as creative potentials, literary devices, cultural context and so on. 
\\2. After thoughtful reflection, clearly indicate your answer by responding YES, NO, or MAYBE. Be honest and precise in your assessment, ensuring each judgment is thoughtfully justified by your analysis.} \\
\midrule
\multicolumn{2}{c}{\textbf{shared part}}\\
\midrule
\multicolumn{2}{l}{\makecell[{{p{7.3cm}}}]{
\textbf{Source text}: \{source\} 
\\\textbf{Translation}: \{translation\} 
\\
\\Please answer YES, NO, or MAYBE to each of the following questions: 
\\\{questions\} 
\\
\\Format your response as a JSON object where each question number is a key and the answer (YES, NO, or MAYBE) is the value. 
\\Do not include explanations, only YES, NO, or MAYBE answers. 
\\Example format: 
\\\{\{ '1': 'YES', '2': 'NO', '3': 'MAYBE' \}\} 
\\
\\Answer: }}\\

\bottomrule
\end{tabular}
\caption{\transpro \va and \pstep templates. Shared parts show texts used in both templates.}\label{tab:template}
\end{table}

\paragraph{Question list} The development of question list consists of three key steps: (1) question collection---gathering diverse literary translation-related questions from textbooks, studies, and practical sources such as blogs and translator interviews, (2) question selection through professional literary translators' votes to identify the most critical and relevant assessment criteria, and (3) question selection through LLM sensitivity analysis to determine which aspects LLMs can effectively assess.
\begin{itemize}\setlength\itemsep{-0.15em}
\item \textbf{Step 1: Question collection.}
We begin by collecting translation-related questions from literary translation research and practices. 
After refinement, we compile 45 questions covering 6 aspects, as shown in Table \ref{tab:qalist1} and \ref{tab:qalist2} (appendix): (1) Grammar \& linguistics, (2) Literary devices, (3) Cultural understanding, context, \& adaptation, (4) Tone \& authorial voice, (5) Consistency \& coherence, and (6) General equivalence. This ensures comprehensive coverage of linguistic, stylistic, and cultural features unique to the literary domain.
\item \textbf{Step 2: Question selection via professional translators’ votes.}
We recruit professional translators to conduct a survey assessing all 45 questions collected previously. Seven professional literary translators (3 male and 4 female) with proof of experience in literary translation from English to other languages (work experience, publications, or educational background) are hired from Upwork.\footnote{\url{https://www.upwork.com/}. \zr{For survey details, see Section \ref{apdx:survay} including screenshots in Figures \ref{fig:screen-ins} \& \ref{fig:screenqa} and distribution of inter-annotation agreements in Figure \ref{fig:agreement} (appendix).}} The translators rate each question on a scale of 5 and give reasons for scores. Surveys take 12 hours in total, averaging 1.7 hours per survey. Translators receive \$12 to \$35 per hour based on experience, totaling \$217.5. We rank questions based on their average ratings and eliminate questions scoring below 4.

\item \textbf{Step 3: Question selection via LLM sensitivity.} We divide the evaluation dataset \liteval and \iyyer into development and test sets as shown in Table \ref{tab:dataset} (see Section \ref{apdx:dev} for details). All test sets remain unseen during the development process. We perform a sensitivity analysis on all 45 questions. Using \va template defined above, we query answers for all 45 questions. We then eliminate questions with low distinguishing power where answer distributions are heavily skewed toward one response, e.g., over 90\% \textit{yes} as shown in Table \ref{tab:qalist1} and \ref{tab:qalist2}.
\end{itemize}

Following the selection process, each question undergoes both professional voting and LLM sensitivity checks independently. Our final list contains 25 questions. We incorporate these into the evaluation prompt by replacing {question} with the 25 questions. Table \ref{tab:qalist1} and Table \ref{tab:qalist2} (appendix) show the complete question list with translator votes and selection status. Our question selection step improves the cost- and compute-efficiency by reducing 20 questions of 451 tokens per query and boosts the metric performance by 0.05 in correlation compared to the unselected list on the development set.

\paragraph{Prompting setup}
We evaluate several prompting templates, as detailed in Table \ref{tab:exp_setup}. The \va template (shown in Table \ref{tab:template}) employs minimal instructions. For more structured approaches, we develop templates with explicit stepwise instructions at two levels: at the entire prompt level (\pstepwei vs.\ \pstep) and at the granular question level (\qstepwei vs.\ \qstep). For \qstep, we craft step-by-step instructed questions, as demonstrated in Table \ref{tab:qstepvsqa} (using \va template). This setup allows us to examine how stepwise instructions and translator-vote weighting influence the evaluation of literary translation quality.

\section{Experiment results}
\begin{table*}\centering
\scriptsize
\begin{tabular}{lcccccccccc}\toprule
\multicolumn{11}{c}{\cellcolor[HTML]{d0e0e3}\textbf{Test set 1: \liteval }} \\\cmidrule{1-11}
\cellcolor[HTML]{d0e0e3}\textbf{Metric} &\multicolumn{2}{c}{\cellcolor[HTML]{d0e0e3}\acceq} &\multicolumn{2}{c}{\cellcolor[HTML]{d0e0e3}\kendall} &\multicolumn{2}{c}{\cellcolor[HTML]{d0e0e3}\makecell{human > top systems \\ (\gptfo, \gtr, \deepl, Qwen)}} &\multicolumn{2}{c}{\cellcolor[HTML]{d0e0e3}human > all systems } &\multicolumn{2}{c}{\cellcolor[HTML]{d0e0e3}\makecell{human > all \\ excluding top systems}} \\\midrule
\multicolumn{11}{c}{\textbf{Human Evaluation}} \\
\textbf{MQM} &\multicolumn{2}{c}{-} &\multicolumn{2}{c}{-} &\multicolumn{2}{c}{\textbf{45.3\%}} &\multicolumn{2}{c}{43.6\%} &\multicolumn{2}{c}{86.8\%} \\
\multicolumn{11}{c}{\textbf{SOTA metrics}} \\
\textbf{\gemba } &\multicolumn{2}{c}{0.534} &\multicolumn{2}{c}{0.561} &\multicolumn{2}{c}{6.1\%} &\multicolumn{2}{c}{6.1\%} &\multicolumn{2}{c}{\textbf{63.1\%}} \\
\textbf{\cometkiwi} &\multicolumn{2}{c}{\textbf{0.552}} &\multicolumn{2}{c}{0.455} &\multicolumn{2}{c}{7.3\%} &\multicolumn{2}{c}{6.2\%} &\multicolumn{2}{c}{52.6\%} \\
\textbf{\cometxl} &\multicolumn{2}{c}{0.528} &\multicolumn{2}{c}{0.387} &\multicolumn{2}{c}{17.0\%} &\multicolumn{2}{c}{12.0\%} &\multicolumn{2}{c}{54.5\%} \\
\textbf{\cometxxl} &\multicolumn{2}{c}{0.540} &\multicolumn{2}{c}{0.400} &\multicolumn{2}{c}{\textbf{26.7\%}} &\multicolumn{2}{c}{\textbf{23.9\%}} &\multicolumn{2}{c}{61.2\%} \\
\textbf{M-Prometheus} &\multicolumn{2}{c}{0.445} &\multicolumn{2}{c}{\textbf{0.570}} &\multicolumn{2}{c}{16.5\%} &\multicolumn{2}{c}{14.8\%} &\multicolumn{2}{c}{56.7\%} \\
\textbf{TREQA-QE} &\multicolumn{2}{c}{0.469} &\multicolumn{2}{c}{0.314} &\multicolumn{2}{c}{12.0\%} &\multicolumn{2}{c}{6.6\%} &\multicolumn{2}{c}{22.0\%} \\
\multicolumn{11}{c}{\textbf{\cometxl Finetuned}} \\
\textbf{\cometxl} &0.528 &$\Delta$ &0.387 &$\Delta$ &17.0\% &$\Delta$ &12.0\% &$\Delta$ &54.5\% &$\Delta$ \\
\textbf{\qmono } &\textbf{0.542} &\textbf{0.014} &0.406 &0.019 &10.3\% &-6.7\% &8.1\% &-3.9\% &49.2\% &-5.3\% \\
\textbf{\hmono } &0.500 &-0.028 &0.394 &0.007 &\textbf{29.4\%} &\textbf{12.4\%} &\textbf{22.3\%} &\textbf{10.3\%} &47.6\% &-6.9\% \\
\textbf{\qmulti } &0.493 &-0.035 &0.348 &-0.039 &23.4\% &6.4\% &21.1\% &9.2\% &42.3\% &-12.1\% \\
\textbf{\hmulti } &0.515 &-0.013 &0.388 &0.001 &20.2\% &3.2\% &18.5\% &6.5\% &50.2\% &-4.2\% \\
\textbf{\textsc{1/4}-WMT24} &0.479 &-0.049 &0.326 &-0.061 &19.9\% &2.9\% &12.7\% &0.8\% &45.0\% &-9.5\% \\
\textbf{\qmono-WMT24} &0.442 &-0.086 &\textbf{0.416} &\textbf{0.029} &12.1\% &-4.9\% &10.4\% &-1.5\% &36.7\% &-17.8\% \\
\textbf{\qmulti-WMT24} &\textbf{0.542} &\textbf{0.014} &0.397 &0.010 &12.9\% &-4.1\% &11.8\% &-0.1\% &\textbf{50.8\%} &\textbf{-3.6\%} \\
\cellcolor[HTML]{efefef}\textbf{Avg.} &\cellcolor[HTML]{efefef}0.502 &\cellcolor[HTML]{efefef}\textcolor{red}{-0.026} &\cellcolor[HTML]{efefef}0.382 &\cellcolor[HTML]{efefef}\textcolor{red}{-0.005} &\cellcolor[HTML]{efefef}18.3\% &\cellcolor[HTML]{efefef}\textcolor{blue}{1.3\%} &\cellcolor[HTML]{efefef}15.0\% &\cellcolor[HTML]{efefef}\textcolor{blue}{3.0\%} &\cellcolor[HTML]{efefef}46.0\% &\cellcolor[HTML]{efefef}\textcolor{red}{-8.5\%} \\
\multicolumn{11}{c}{\textbf{\transpro}} \\
\textbf{BEST SOTA} &0.552 &$\Delta$ &0.570 &$\Delta$ &26.7\% &$\Delta$ &23.9\% &$\Delta$ &63.1\% &$\Delta$ \\
\textbf{\va$\dagger$} &0.606 &0.054 &\textbf{0.605} &\textbf{0.035} &38.7\% &12.0\% &37.0\% &13.1\% &\textbf{85.7\%} &\textbf{22.5\%} \\
\textbf{\vawei$\dagger$} &\textbf{0.616} &\textbf{0.063} &\textbf{0.605} &\textbf{0.035} &\textbf{41.4\%} &\textbf{14.7\%} &\textbf{40.3\%} &\textbf{16.4\%} &\textbf{85.7\%} &\textbf{22.5\%} \\
\textbf{\pstep$\dagger$} &0.585 &0.033 &0.585 &0.015 &31.9\% &5.3\% &29.7\% &5.8\% &82.3\% &19.2\% \\
\textbf{\pstepwei$\dagger$} &0.594 &0.042 &0.587 &0.017 &36.3\% &9.6\% &34.1\% &10.2\% &84.0\% &20.9\% \\
\textbf{\qstep$\dagger$} &0.595 &0.043 &0.594 &0.024 &25.9\% &-0.8\% &22.5\% &-1.4\% &80.1\% &17.0\% \\
\textbf{\qstepwei$\dagger$} &0.600 &0.048 &0.597 &0.027 &27.0\% &0.3\% &23.7\% &-0.2\% &80.7\% &17.6\% \\
\cellcolor[HTML]{efefef}\textbf{Avg.} &\cellcolor[HTML]{efefef}0.599 &\cellcolor[HTML]{efefef}\textcolor{blue}{0.047} &\cellcolor[HTML]{efefef}0.595 &\cellcolor[HTML]{efefef}\textcolor{blue}{0.026} &\cellcolor[HTML]{efefef}33.5\% &\cellcolor[HTML]{efefef}\textcolor{blue}{6.9\%} &\cellcolor[HTML]{efefef}31.2\% &\cellcolor[HTML]{efefef}\textcolor{blue}{7.3\%} &\cellcolor[HTML]{efefef}83.1\% &\cellcolor[HTML]{efefef}\textcolor{blue}{20.0\%} \\
\bottomrule
\end{tabular}
\caption{Results for \liteval. \acceq and \kendall measure the segment-level correlation between human  MQM and metrics. $\Delta$ indicates changes in absolute value. The metric adequacy is reported as the percentage of cases where the best human translation is scored higher than outputs from (1) top systems, (2) all systems, and (3) all systems but top. 
$\dagger$ indicates significantly better \acceq and \kendall compared to the best SOTA on at least 3 out of 4 language pairs with permutation test at $p < 0.05$.}\label{tab:res_liteval}
\end{table*}
We evaluate metric performance on 3 datasets: \liteval in Table \ref{tab:res_liteval}, \iyyer in Table \ref{tab:res_iyyer} and \paran in Table \ref{tab:res_par3} (appendix). We analyze metrics' correlation with human judgments and their adequacy, i.e., the ability to rank human translations over MT outputs. We include SOTA metrics as baselines: finetuned metrics (\cometxl, \cometxxl, and \cometkiwi), LLM-based prompting metric \gemba, recent QA metric TREQA, and recent multilingual LLM M-Prometheus (14b), trained for general evaluation purposes \cite{pombal2025mprometheussuiteopenmultilingual}. We compare literary-finetuned \cometxl against the original \cometxl and compare \transpro against the best available metric performance. To ensure comparability, all LLM-based prompting metrics use the same base model (GPT-4o-mini). We also show \transpro performance on other base models in Table \ref{tab:model_stab} and \ref{tab:res_par3} (appendix).

\subsection{Marginal gains for finetuning \cometxl}
Finetuning \cometxl on literary tasks offers modest improvements in some settings, though the benefits are inconsistent. 
For \liteval, finetuned metrics like \qmono and \hmono show mild performance gains over the base \cometxl with \qmono's \acceq from 0.528 to 0.542 and \kendall from 0.387 to 0.406. However, these improvements do not consistently translate to adequacy gains. Other finetuned metrics like \textsc{1/4-WMT24} and \hmulti show mixed results. Resource-intensive configurations even degrade performance---\qmulti shows a 12.1-point drop in \textit{Human > all but top systems}. Joint finetuning with both tasks barely improves the performance. On average, finetuned methods show mixed results with correlation slightly below \cometxl and minor adequacy gains with 1-3 points for 2 testing cases \textit{Human > top \& Human > all}. For \iyyer and \paran, the impact of finetuning is more negative. \qmulti-WMT24 leads among the finetuned versions for \iyyer, with \acceq reaching 0.643 and \kendall 0.292. However, its adequacy drops from 18.5\% to 14.4\%. While \hmono and \qmulti show modest improvements in adequacy score by 2 points to 20.5\%, other finetuned variants show barely any improvements.

Our analysis suggests that finetuning yields marginal gains with bilingual datasets or shallow layers being more effective. This echoes the finding from \citet{Shilongtail} where focused tuning on shallow layers achieves better alignment with literary translation goals by concentrating on essential features instead of noise from inconsistent literary styles in multilingual datasets.
\begin{table}[ht]
\centering
\tiny
\begin{tabular}{lcccccc}\toprule
\multicolumn{7}{c}{\cellcolor[HTML]{d9d2e9}\textbf{Test set 2: \iyyer}} \\\cmidrule{1-7}
\cellcolor[HTML]{d9d2e9}\textbf{Metric} &\multicolumn{2}{c}{\cellcolor[HTML]{d9d2e9}\acceq} &\multicolumn{2}{c}{\cellcolor[HTML]{d9d2e9}\kendall} &\multicolumn{2}{c}{\cellcolor[HTML]{d9d2e9}\makecell{human > MT \\ (\gpttf \& \gtr)}} \\\midrule
\multicolumn{7}{c}{\textbf{SOTA metrics}} \\
\textbf{\gemba} &\multicolumn{2}{c}{0.419} &\multicolumn{2}{c}{\textbf{0.269}} &\multicolumn{2}{c}{11.8\%} \\
\textbf{\cometkiwi} &\multicolumn{2}{c}{0.586} &\multicolumn{2}{c}{0.172} &\multicolumn{2}{c}{9.6\%} \\
\textbf{\cometxl} &\multicolumn{2}{c}{\textbf{0.603}} &\multicolumn{2}{c}{0.207} &\multicolumn{2}{c}{18.5\%} \\
\textbf{\cometxxl} &\multicolumn{2}{c}{0.586} &\multicolumn{2}{c}{0.171} &\multicolumn{2}{c}{\textbf{26.0\%}} \\
\textbf{M-Prometheus} &\multicolumn{2}{c}{0.223	} &\multicolumn{2}{c}{0.124	} &\multicolumn{2}{c}{21.8\%} \\
\textbf{TREQA-QE} &\multicolumn{2}{c}{0.519} &\multicolumn{2}{c}{0.038} &\multicolumn{2}{c}{14.4\%} \\

\multicolumn{7}{c}{\textbf{\cometxl Finetuned}} \\
\textbf{\cometxl} &0.603 &$\Delta$ &0.207 &$\Delta$ &18.5\% &$\Delta$ \\
\textbf{\qmono } &0.637 &0.034 &0.275 &0.068 &8.3\% &-10.3\% \\
\textbf{\hmono } &0.583 &-0.021 &0.183 &-0.024 &\textbf{20.5\%} &\textbf{2.0\%} \\
\textbf{\qmulti } &0.574 &-0.029 &0.148 &-0.059 &20.4\% &1.9\% \\
\textbf{\hmulti } &0.567 &-0.036 &0.188 &-0.019 &12.4\% &-6.1\% \\
\textbf{\textsc{1/4}-WMT24} &0.576 &-0.027 &0.153 &-0.054 &15.7\% &-2.8\% \\
\textbf{\qmono-WMT24} &0.474 &-0.130 &\textbf{0.312} &\textbf{0.105} &8.9\% &-9.6\% \\
\textbf{\qmulti-WMT24} &\textbf{0.643} &\textbf{0.040} &0.292 &0.086 &14.4\% &-4.1\% \\
\cellcolor[HTML]{efefef}\textbf{Avg.} &\cellcolor[HTML]{efefef}0.579 &\cellcolor[HTML]{efefef}\textcolor{red}{-0.024} &\cellcolor[HTML]{efefef}0.222 &\cellcolor[HTML]{efefef}\textcolor{blue}{0.015} &\cellcolor[HTML]{efefef}14.4\% &\cellcolor[HTML]{efefef}\textcolor{red}{-4.1\%} \\
\multicolumn{7}{c}{\textbf{\transpro}} \\
\textbf{BEST SOTA} &0.603 &$\Delta$ &0.269 &$\Delta$ &26.0\% &$\Delta$ \\
\textbf{\va} &0.519 &-0.085 &0.303 &0.035 &38.9\% &12.9\% \\
\textbf{\vawei} &\textbf{0.570} &\textbf{-0.033} &\textbf{0.304} &\textbf{0.035} &\textbf{42.4\%} &\textbf{16.4\%} \\
\textbf{\pstep} &0.466 &-0.138 &0.258 &-0.010 &36.3\% &10.3\% \\
\textbf{\pstepwei} &0.510 &-0.093 &0.278 &0.009 &39.1\% &13.1\% \\
\textbf{\qstep} &0.528 &-0.075 &0.290 &0.021 &33.4\% &7.4\% \\
\textbf{\qstepwei} &0.554 &-0.049 &0.291 &0.023 &35.5\% &9.5\% \\
\cellcolor[HTML]{efefef}\textbf{Avg.} &\cellcolor[HTML]{efefef}0.525 &\cellcolor[HTML]{efefef}\textcolor{red}{-0.079} &\cellcolor[HTML]{efefef}0.287 &\cellcolor[HTML]{efefef}\textcolor{blue}{0.019} &\cellcolor[HTML]{efefef}37.6\% &\cellcolor[HTML]{efefef}\textcolor{blue}{11.6\%} \\
\bottomrule
\end{tabular}
\caption{Results for \iyyer. \acceq and \kendall measure the segment-level correlation between human judgments (pairwise preference) and metrics. $\Delta$ indicates changes in absolute value. The metric adequacy is reported as the percentage of cases where human translation is scored higher than the outputs from \gpttf and \gtran (\gtr).}\label{tab:res_iyyer}
\end{table}

\subsection{\transpro: strong performance gain in correlation and adequacy}
\transpro demonstrates substantial and consistent improvements in both correlation with human judgments and adequacy (see example in Tables \ref{tab:qualitative}). For \liteval, \transpro outperforms the best SOTA baselines by a large margin. All variants of \transpro show significantly better correlation than the best SOTA as shown in Table \ref{tab:res_liteval}. \vawei shows the strongest performance with a score of 0.616 for \acceq and 0.605 for \kendall. It also demonstrates the highest gains for all adequacy cases, with \textit{Human > top systems} at 41.4\%, \textit{Human > all systems} at 40.3\%, \textit{Human > all systems but top} at 85.7\%---a 14.7, 16.4, and 22.5 point increase over the best SOTA. \transpro outperforms the best finetuned \cometxl, showing an increase of nearly 0.2 in \kendall, 18-point and 35-point in adequacy for \textit{Human > all systems} and \textit{Human > all but top systems}. For \iyyer in Table \ref{tab:res_iyyer}, \transpro continues to excel, delivering the strongest adequacy results while maintaining top-level correlation. \vawei achieves a \kendall of 0.304 and an adequacy of 42.4\%, marking a 16.4-point gain over the SOTA \cometxxl. For \paran, \transpro again outperforms best SOTA metrics by 0.06 in \kendall and nearly 18 points in adequacy on average. Also worth noting is that step instructions, both \pstep and \qstep, perform worse than \va setting. This may suggest that detailed, literary-specific instructions, particularly \qstep, could dilute the effectiveness of LLMs' judgments in complex tasks compared to simpler instructions (see example in Tables \ref{tab:qualitative_f}).  

\paragraph{Translator votes improve metric performance.}
Translator vote-weighted variants improve correlation and adequacy compared to unweighted scores, as shown in Table \ref{tab:translator_votes}, Figure \ref{fig:comparison}, \zr{and significance analysis in Section \ref{apdx:improve_weights} (appendix)}. For \liteval, the weighted versions achieve a better score by nearly 0.01 in \acceq and over 4 points in adequacy compared to their unweighted versions. For \iyyer, weighting with translator votes is a key differentiator, with all evaluation cases showing improvements: 0.05 in \acceq, 0.02 in \kendall, and 3.5 points in adequacy. \paran shows similar gains in \acceq and adequacy. Overall, weighted scores yield average improvements of 0.02 in \acceq and 2 points in adequacy. These results demonstrate the value of incorporating professional translators' perspectives into the evaluation.

\paragraph{\transpro achieves high adequacy, approaching student annotator performance.} For \liteval, \transpro's adequacy results closely reach student annotator performance across all three comparison cases, with the gap narrowing to less than 4 points for all cases.\footnote{\zr{Our adequacy results differ from those reported in \citet{zhang2024good} for two reasons: (1) we partition the dataset into test and development sets and report results only on the test set, while also making these splits available for reproducibility; and (2) Table 3 in \citet{zhang2024good} uses only a subset of the full data (see their footnote 12 regarding the BWS samples).}} This marks a substantial improvement over existing SOTAs, which show gaps of 18.6 points for \textit{human > top systems}, 19.7 points for \textit{human > all systems}, and 23.7 points for \textit{human > all but top systems}. While \transpro has not yet matched the professional human translators' performance of nearly 90\% for \textit{human > top systems} \cite{zhang2024good}, it demonstrates noticeable progress toward human-level evaluation capabilities \zr{comparable to trained linguistics students}, highlighting its potential \zr{and room for future improvements} for literary MT evaluation.

\paragraph{Can \transpro perform well using other base models?} To evaluate \transpro's compatibility on other base models, we implement the \pstep on open-source models of different sizes. Table \ref{tab:model_stab} and \ref{tab:res_par3} (appendix) indicate that open-source models \llama3.3-70b and Qwen2.5-32b show competitive results for all datasets. For \liteval and \iyyer, open-source models achieve even better results than GPT-4o-mini with \llama3.3-70b surpassing GPT-4o-mini in adequacy by 9 points on \textit{human > top systems} and Qwen2.5-32b exceeding by over 0.1 \acceq and 0.07 \kendall for \iyyer. \llama-70b's reasoning variant DeepSeek-distilled shows unsatisfactory performance. Additionally, \llama's previous larger version 405b lags behind the smaller 70b model. Our results indicate that \transpro maintains competitiveness when applied to open-source models, demonstrating its generalizability.

\subsection{Ablation study for \transpro}
\begin{table}[!htp]\centering
\centering
\tiny
\begin{tabular}{ccccc}\toprule
\multicolumn{5}{c}{\cellcolor[HTML]{d9d2e9}\textbf{Test set 2: \iyyer}}\\
\cellcolor[HTML]{d9d2e9}\textbf{Metric/Asp.} &\cellcolor[HTML]{d9d2e9}\textbf{\#Qs} &\cellcolor[HTML]{d9d2e9}\acceq &\cellcolor[HTML]{d9d2e9}\kendall &\cellcolor[HTML]{d9d2e9}\makecell{human > LLM \\ (\gpttf \& \gtr)} \\\midrule
\textbf{\va} &25 &0.519 &0.303 &38.9\% \\
\multicolumn{5}{c}{\textbf{ablation setting 1: scores with one aspect alone (+)}} \\
\textbf{GL} &3 &0.171 &0.177 &11.0\% \\
\textbf{LD} &4 &0.231 &0.199 &21.3\% \\
\textbf{CCA} &5 &0.285 &0.210 &24.5\% \\
\textbf{TA} &6 &0.327 &0.282 &21.8\% \\
\textbf{CO} &2 &0.033 &0.042 &0.0\% \\
\textbf{GE} &5 &0.345 &0.276 &26.9\% \\
\multicolumn{5}{c}{\textbf{ablation setting 2: scores excluding one aspect (-)}} \\
\textbf{GL} &22 &0.512 &0.304 &36.9\% \\
\textbf{LD} &21 &0.504 &0.311 &39.7\% \\
\textbf{CCA} &20 &0.483 &0.301 &36.3\% \\
\textbf{TA} &19 &0.428 &0.260 &36.3\% \\
\textbf{CO} &23 &0.516 &0.304 &38.3\% \\
\textbf{GE} &20 &0.488 &0.295 &32.8\% \\
\bottomrule
\end{tabular}
\caption{Ablation results per question aspects on \iyyer. Asp. stands for aspects: (1) GL (Grammar \& linguistics); (2) LD (Literary devices); (3) CCA (Cultural understanding, context, \& adaptation); (4) TA (Tone \& authorial voice); (5) CO (Consistency \& coherence); and (6) GE (General equivalence). \va represents \transpro score using the full selected question set. \#Q indicates the number of questions.}\label{tab:ablation}
\end{table}
\zr{We conduct an ablation study to investigate the contribution of each literary aspect of \transpro. Table \ref{tab:ablation} reports the results on \iyyer dataset under two ablation settings: (1) scoring with one aspect alone; (2) scoring excluding one particular aspect. We draw three key conclusions. First, the contribution of \textit{any single aspect in isolation} (setting 1) is substantially weaker than the full question set (\va). Even the strongest single aspect, GE (General equivalence), lags behind \va by more than 0.17 in \acceq and 12 points in adequacy. Second, \textit{removing any single aspect} (setting 2) does not yield improvements over \va across all three evaluation perspectives. While excluding LD (Literary devices) leads to slight gains in \kendall and adequacy, it incurs a non-negligible drop in \acceq by 0.015, while the removal of other aspects consistently produces declines. This demonstrates that \transpro benefits from the complementary contributions of all six aspects. Third, some dimensions are more impactful than others. For example, in setting 2, excluding  TA (Tone \& authorial voice) causes the largest degradation in \acceq by 0.09, while omitting GE results in the largest decline in adequacy, highlighting their central roles. In contrast, removing CO (Consistency \& coherence) has comparatively limited effect.} 
\section{Conclusion}
In this paper, we introduce \transpro, a novel LLM-based QA metric specifically designed for evaluating literary translations. \transpro addresses critical shortcomings of existing approaches and achieves substantial performance gains over both finetuned \cometxl and current SOTA metrics. Our results show improvements across all 3 test sets, with increases of 0.04 in \kendall, 0.06 in \acceq, and over 15 points in adequacy. \transpro's strong performance with open-source models enhances its accessibility and reduces dependency on proprietary technology, while enabling broader adoption and careful consideration of ethical issues in evaluating copyrighted and culturally sensitive texts.

\transpro is designed to better account for aspects of translation quality emphasized in professional human translation, such as creative subtleties and cultural nuances often overlooked by literal and homogenized MT outputs. In doing so, \transpro can help mitigate the existing bias toward literal translations and support the recalibration of LLMs toward more human-like literary translation.

\section*{Limitations}
While we try to cover as many languages as possible, our evaluation remains predominantly focused on high- and medium-resource language pairs due to the limited availability of suitable evaluation datasets. This underscores the necessity of developing comprehensive evaluation datasets within the literary domain, particularly targeting low-resource language pairs. \zr{Future work can also explore the effect of genre and style variation on the metric. While our current analysis mainly relies on the opinions of literary translators, incorporating feedback from broader audiences could provide additional insight.}

Additionally, we currently evaluate translations at the paragraph level, which may miss subtle literary elements that span larger narrative sections. A key limitation is the absence of evaluation datasets containing extended narrative units like consecutive chapters or complete works. Future research could expand the evaluation dataset to include wider contexts.

Our experiments show that question-level instruction templates perform less effectively than the two simpler configurations. While we hypothesize this stems from LLMs' lack of specialized literary knowledge, our current study does not analyze the specific nature or extent of this knowledge gap. Further investigation into LLMs' specialized knowledge of literary and creative tasks remains a valuable research direction. 

Finally, although \transpro performs substantially well on both closed- and open-source models, it could benefit further from domain-specific finetuning of LLMs \cite{rafailov2023direct}. Future research can delve into even smaller models using this method to improve efficiency. 
\section*{Ethical Considerations}
We utilize open-source datasets for evaluation and finetuning. For datasets containing copyrighted content, we use them following fair use principles for research and academic purposes.  

For human evaluation, we obtained informed consent from all participating professional translators. Their contributions are disclosed anonymously and do not include any protected demographic or personal information. 

\paragraph{Potential risks}
\zr{Potential risks of \transpro include reinforcing biases toward high- and medium-resource languages, while effects remain unknown for underrepresented low-resource languages in test data. To ensure equitable distribution of \transpro's benefits, these risks demand careful attention through responsible deployment and more comprehensive dataset coverage.} 

\paragraph{Licensing and intended use}
\zr{Our implementation builds on components from GEMBA-MQM (CC-BY-SA-4.0), M-Prometheus (Apache-2.0), and COMET (Apache-2.0). Because GEMBA-MQM is licensed under CC-BY-SA-4.0, which includes a ShareAlike clause, our released code is distributed under the same license. By contrast, Apache-2.0 components are permissive and compatible with redistribution under CC-BY-SA-4.0. Our use of these artifacts, as well as the LLMs utilized in this work, complies with their respective licenses and use policies. This release is intended solely for research and evaluation purposes in literary translation.}

\paragraph{PII in data}
\zr{We rely exclusively on existing publicly available datasets. We have not performed independent, systematic checks for personally identifiable information (PII) within these datasets, as we consider this the responsibility of the original dataset creators. As our work involves literary excerpts, some texts may contain offensive language. In literary contexts, the presence of such language is not inherently undesirable, as it reflects the source material and, in some cases, is directly relevant to the research question.}

\paragraph{Packages}
\zr{We use the mt-metrics-eval package (version 2, commit 6d4b0bb), with a modification to meta\_info.py to include our test dataset meta information. The following important dependencies are used: scipy 1.10.1, seaborn 0.13.2, transformers 4.50.0, and openai 1.68.2 (for API calls).}
\section*{Acknowledgements}
We thank the anonymous reviewers for their feedback, which greatly improved the work. We appreciate the professional input from all professional translators involved. The NLLG Lab gratefully acknowledges support from the Federal Ministry of Education and Research (BMBF) via the research grant ``Metrics4NLG'' and the German Research Foundation (DFG) via the Heisenberg Grant EG 375/5-1.

\bibliography{acl_latex}

\appendix
\section{Appendix}\label{sec:appdx}
\subsection{Datasets}\label{apdx:dataset}
Table \ref{tab:dataset} summarizes the statistics of both evaluation and finetuning datasets. 
\begin{table*}\centering
\scriptsize
\begin{tabular}{lcccccccc}\toprule
\multirow{2}{*}{\textbf{Dataset}} &\multirow{2}{*}{\textbf{Use case}} &\multirow{2}{*}{\textbf{Language Pair}} &\multirow{2}{*}{\textbf{Book}} &\multicolumn{2}{c}{\textbf{Size}} &\multirow{2}{*}{\textbf{\#MT systems}} & \multirow{2}{*}{\textbf{Annotation Type} }  \\\cmidrule{5-6}
& & & &\textbf{test} &\textbf{dev} & \\\midrule
\textbf{\liteval} &\multirow{2}{*}{\textbf{Test-Dev}} &\deen, \enDe, \dezh, \enzh &Contemporary/Classics &1996 &70 &9 & MQM score\\
\textbf{\iyyer} & &\makecell{Src: En, De, Fr, Ru, Cs, Ja \\ Tgt: En, Ja, Pl } &Contemporary &1095 &165 &2 & Preference \\
\textbf{\paran} & \textbf{Test} & Fr-En, Ru-En, De-En & Classics & \multicolumn{2}{c}{744} & 2 & Preference\\
\hdashline
\textbf{\parun} 
&\makecell{\textbf{Finetuning} \\ (ranking)} &\makecell{Src: 18 languages \\ Tgt: En} &Classics &\multicolumn{2}{c}{50k} & 6 & - \\
\textbf{WMT24} 
&\makecell{\textbf{Finetuning} \\ (regression)} & En-Cs, Hi, Is, Ja, Ru, Uk, Zh & Contemporary &\multicolumn{2}{c}{4600} & 8-13 & ESA score \\
\bottomrule
\end{tabular}
\caption{Summary statistics of evaluation and finetuning datasets. Size indicates the number of source-target pairs of paragraphs. The ESA score, i.e., error span annotation \cite{kocmi-etal-2024-error} is an updated version of MQM (Multidimensional Quality Metrics). Preference refers to direct preference comparison between pairs of translation versions, without assigning numerical scores.}\label{tab:dataset}
\end{table*}
\paragraph{\liteval} combines sources from contemporary and classic literary texts, including translations from 9 MT systems: the GPT series (\gptfo), commercial models (\gtran and \deepl), popular smaller LLMs (\llama3, \qwen 2, Gemini, \tower), and previous SOTA systems (M2M, \nllb). Link to the official WMT package: \url{https://github.com/google-research/mt-metrics-eval}. \acceq is a variant of \kendall that is recently proposed and implemented by the WMT shared task \cite{deutsch-etal-2023-ties}. This metric evaluates pairwise accuracy while accounting for tie calibration. We report both scores to ensure broader comparability. 

\paragraph{\iyyer} Three prompting methods are examined for translating paragraphs using \gpttf: translating sentence-by-sentence without context (\sent), translating sentence-by-sentence with full paragraph context (\sentp), and directly translating a whole paragraph (\para). The dataset includes direct pairwise preference annotations comparing \sent vs.\ \para, \sentp vs.\ \para, and \gtran vs.\ \para.

\paragraph{Adequacy} For the adequacy measure of cases with multiple human translations, we consider the version rated highest in human evaluations. This approach is reasonable since older translations may be less appealing to modern annotators due to changes in language and style over time. \zr{Regarding the adequacy performance of annotators (human level), only \liteval provides annotations scoring both human translation and MT outputs. In contrast, \iyyer and PAR3 only contain human annotations of MT outputs, making it impossible to determine annotators' adequacy performance.}

\paragraph{Correlation metric} \zr{The choice of correlation metrics (\acceq and \kendall) is motivated by two reasons. First, both have been recently adopted in the WMT shared task, one of the most renowned venues for MT evaluation, ensuring comparability with prior work. Second, the \iyyer and PAR3 datasets contain only pairwise comparison data. In this setting, \kendall and \acceq are particularly well-suited, as the number of concordant and discordant pairs is well-defined even under binary labels. By contrast, Pearson and Spearman correlations require continuous or ordinal scores, making them less appropriate for pairwise judgments.}

\subsection{Finetuned \cometxl}
\subsubsection{Example of finetuning dataset}
Table \ref{tab:par3pair} shows an example of a \parun paired dataset for the ranking task.
\begin{table}[H]\centering
\scriptsize
\begin{tabular}{ m{0.65cm} m{5cm}m{1cm}} \toprule
\textbf{Pair} &\textbf{\deen} &\textbf{Model} \\\midrule
\textbf{Source} &Am wiederholtesten aber fragte der treue Diener, fast so oft er Ottilien sah, nach der Rückkunft des Herrn und nach dem Termin derselben. &Human \\
\textbf{Positive} &But almost every time the faithful servant saw Ottilie what he most repeatedly asked about was the master’s return and when that was going to happen. &Human \\
\textbf{Negative} &Most frequently, however, the faithful servant asked, almost every time he saw Ottilie, about the return of his master and the date of that return. &GPT-4o-mini \\
\bottomrule
\end{tabular}
\caption{Example of PAR3 paired dataset for the ranking task.}\label{tab:par3pair}
\end{table}
\subsubsection{Finetuning details}
\begin{table}[H]\centering
\scriptsize
\begin{tabular}{lr}\toprule
parameter & value \\
\midrule
batch\_size &8 \\
encoder\_learning\_rate &2.00e-5 \\
encoder\_weight\_decay &0.01 \\
max\_length &512 \\
gradient\_accumulation\_steps &4 \\
early\_stopping &true \\
epoch &3 \\
loss &Triplet loss \\
\bottomrule
\end{tabular}
\caption{Finetuning parameters}\label{tab:param}
\end{table}
Table \ref{tab:param} shows the finetuning parameters for \cometxl. We use the PyTorch implementation for both losses. See \url{https://pytorch.org/docs/stable/generated/torch.nn.TripletMarginLoss.html} for the formula of triplet loss.   

\subsubsection{Discussion on \cometxl}
\zr{The limited gains from finetuned \cometxl likely stem from multiple factors: (1) Domain mismatch, XCOMET is initially trained on sentence-level non-literary data, while our fine-tuning involved paragraph-level literary domain; (2) Limited training data — regression datasets are rather small and both regression/ranking datasets fail to cover all languages in the test sets; (3) Data quality — genre variation and uncertain translation quality in some datasets, as also noted in WMT 2024, may limit effective adaptation; (4) Literary texts are inherently difficult to learn, especially at paragraph level.}

\subsection{Details for \transpro}\label{apdx:survay}
\subsubsection{Development set details}\label{apdx:dev}
We sample 1-2 source paragraphs per language pair from each dataset. Our development set contains 70 source-target pairs (3.4\%) on 4 language pairs from \liteval and 165 (13.1\%) on 18 language pairs from \iyyer. We make sure that all test sets remain unseen during the development process. Our metric performance is reported on test data only.
\subsubsection{Survey details}
\begin{figure}
    \centering
       \includegraphics[width=0.5\textwidth]{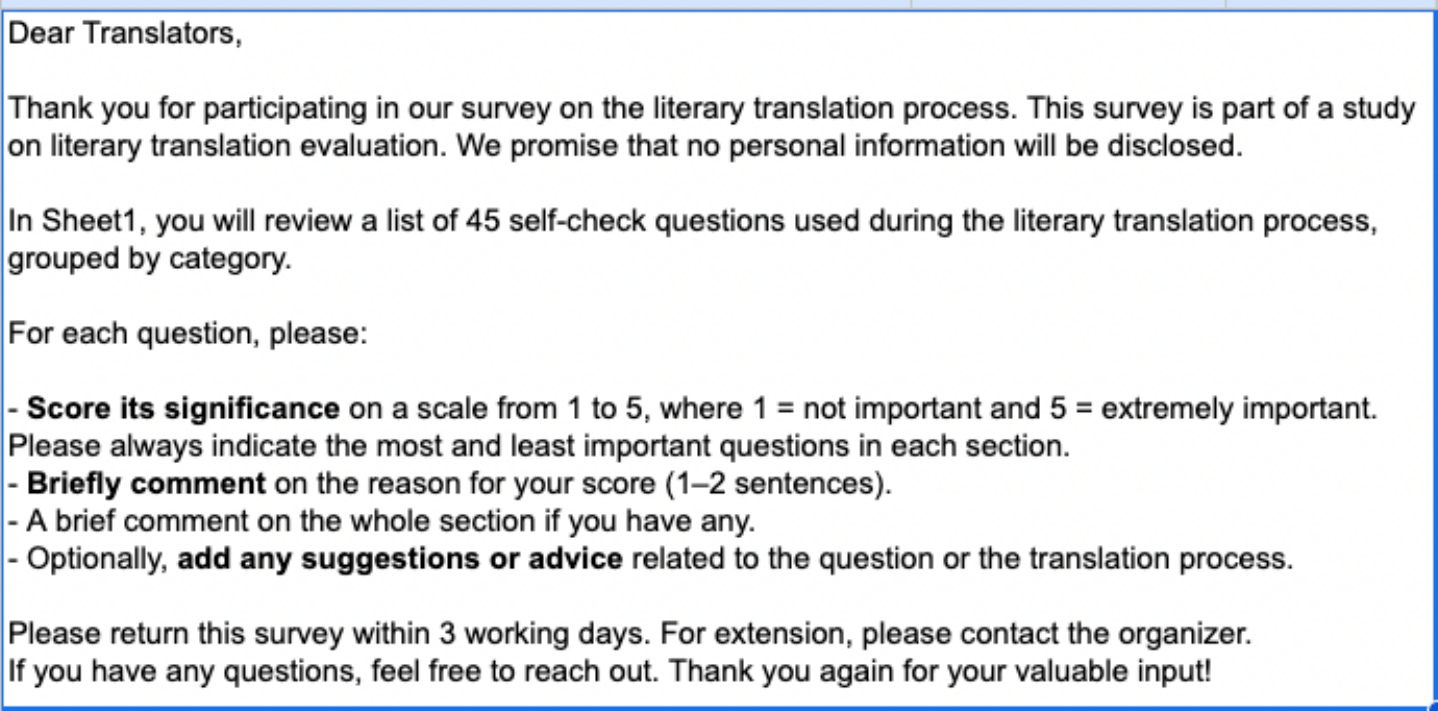}
    \caption{Screenshot of the instruction page for the survey.}
    \label{fig:screen-ins}
\end{figure}

\paragraph{The instruction}
Figure \ref{fig:screen-ins} shows the screenshot of the instruction page for the survey. 
\paragraph{The full question list}
Figure \ref{fig:screenqa} demonstrates the screenshot of the survey page with the complete question list. 

\subsubsection{The selected question list with translator votes}
Tables \ref{tab:qalist1} and \ref{tab:qalist2} present the complete question list with their status, mean translator voting, and reasons for exclusion from the final list. The status indicates one of four outcomes: (1) S for selected questions, (2) R-GI for rejected questions due to general insensitivity in GPT-4o-mini results (where one answer dominated across translations, showing poor general quality discrimination), (3) R-HI for rejected questions due to human insensitivity (where a high percentage of no/maybe responses for human translations indicated poor discrimination of human translation), and (4) R-TV for questions rejected due to low translator vote scores. We highlight the selected questions in \textcolor{YellowGreen}{\textbf{light green}}.

\subsubsection{\zr{Annotation agreement among translators}}
\begin{figure}
    \centering
    \includegraphics[width=0.95\linewidth]{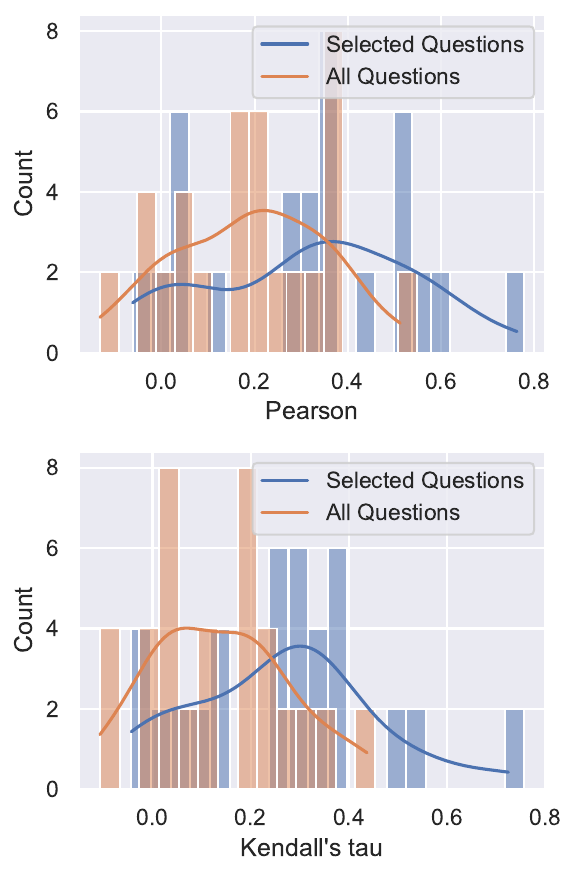}
    \caption{Pairwise inter-annotator agreement distribution for question importance ratings among the seven translators.}
    \label{fig:agreement}
\end{figure}
\zr{As literary translation is a nuanced and inherently subjective task, some disagreement among expert annotators is expected. To obtain a more reliable consensus, we employ seven independent annotators and report pairwise agreement using Pearson and Kendall coefficients. We report the distribution of inter-annotator agreement among the seven translators (all 45 questions vs.\ selected 25 questions) in Figure \ref{fig:agreement}. Rather than presenting averaged values, we show the full distribution because correlation measures are non-additive and their averages may be misleading and hard to interpret \cite{alexander1990note, ji-etal-2022-achieving}.}

\zr{The distributions in Figure \ref{fig:agreement} demonstrate moderate inter-annotator correlations, with centers around 0.3-0.4 on both measures for selected questions. This shows reasonable consistency considering the task's subjective nature. Additionally, selected questions show higher agreement compared to all questions, indicating that the selected subset contains questions with more reliable cross-translator judgments.}

\begin{table}\centering
\tiny
\begin{tabular}{m{1.6cm}m{0.3cm}m{0.5cm}m{0.6cm}m{1cm}m{1cm}}\toprule
\multicolumn{6}{c}{\cellcolor[HTML]{d0e0e3}\liteval} \\\cmidrule{1-6}
Base &reason &\acceq &\kendall &\makecell{human > \\ top systems} &\makecell{human > \\ all systems} \\\midrule
GPT-4o-mini$_w$ & No &0.594 &\textbf{0.587} &36.3\% &\textbf{84.0\%} \\
\hdashline
LLaMa3.1-405b$_w$ & No &0.537 &0.506 &36.5\% &70.0\% \\
LLaMa3.3-70b$_w$ & No &0.552 &0.537 &\textbf{45.0\%} &76.0\% \\
DK LLaMa-70b$_w$ &Yes &0.497 &0.461 &12.6\% &40.9\% \\
Qwen2.5-32b$_w$ &No & \textbf{0.602}	& 0.584	& 31.3\% & 82.9\% \\

\midrule
\multicolumn{6}{c}{\cellcolor[HTML]{d9d2e9}\iyyer} \\
\cmidrule{1-6}
Base &reason &\acceq &\kendall &\multicolumn{2}{c}{\makecell{human > MT \\ (\gpttf \& \gtr)}} \\
\midrule
GPT-4o-mini$_w$ & No &0.510 &0.278 &\multicolumn{2}{c}{\textbf{39.1\%}} \\
\hdashline
LLaMa3.1-405b$_w$ & No &0.385 &0.154 & \multicolumn{2}{c}{22.0\%}  \\
LLaMa3.3-70b$_w$ & No&0.444 &0.208 &\multicolumn{2}{c}{36.8\%} \\
DK LLaMa-70b$_w$ & Yes&0.405 &0.069 &\multicolumn{2}{c}{15.0\%} \\
Qwen2.5-32b$_w$ & No & \textbf{0.616}	& \textbf{0.346}	&\multicolumn{2}{c}{35.7\%} \\
\bottomrule
\end{tabular}
\caption{\transpro performance on open-source base models using \pstep template weighted by translator votes. Reason denotes whether the model has reasoning capabilities. DK stands for DeepSeek distilled version.}\label{tab:model_stab}
\end{table}
\begin{table}\centering
\tiny
\begin{tabular}{lccccc}\toprule
\multicolumn{6}{c}{\cellcolor[HTML]{d0e0e3}\liteval} \\
\cmidrule{1-6}
template&\acceq &\kendall &\makecell{human > \\ top systems} & \makecell{human > \\ all systems} &\makecell{human > \\ all but top} \\\midrule
\textbf{\va} &0.009 &0.000 &2.7\% &3.3\% &0.0\% \\
\textbf{\pstep} &0.009 &0.002 &4.4\% &4.4\% &1.7\% \\
\textbf{\qstep} &0.004 &0.002 &1.1\% &1.1\% &0.6\% \\
\midrule
\multicolumn{6}{c}{\cellcolor[HTML]{d9d2e9}\iyyer} \\
\cmidrule{1-6}
template &\acceq &\kendall &\multicolumn{3}{c}{\makecell{human > MT \\ (\gpttf \& \gtr)}} \\
\midrule
\textbf{\va} &0.052 &0.001 &\multicolumn{3}{c}{3.5\%} \\
\textbf{\pstep} &0.044 &0.019 &\multicolumn{3}{c}{2.8\%} \\
\textbf{\qstep} &0.026 &0.001 &\multicolumn{3}{c}{2.1\%} \\
\midrule
\multicolumn{6}{c}{\cellcolor[HTML]{c9daf8}\textbf{Test set 3: PAR3 annotated}} \\
\cmidrule{1-6}
template &\acceq &\kendall &\multicolumn{3}{c}{\makecell{human > MT \\ (\gpttf \& \gtr)}} \\
\midrule
\textbf{\va} &0.001 &-0.013 &\multicolumn{3}{c}{3.3\%} \\
\textbf{\pstep} &-0.002 &-0.020 &\multicolumn{3}{c}{0.0\%} \\
\midrule
\cellcolor[HTML]{f3f3f3}Avg. &\cellcolor[HTML]{f3f3f3}0.018 &\cellcolor[HTML]{f3f3f3}-0.001 &\multicolumn{3}{c}{\cellcolor[HTML]{f3f3f3}2.1\%} \\
\bottomrule
\end{tabular}
\caption{Impact of translator-weighted scores for \transpro across 3 evaluation sets. The table shows \transpro's absolute performance gains ($\Delta$) when using weighted versus non-weighted scoring. Avg. represents the mean across all datasets.}\label{tab:translator_votes}
\end{table}

\subsubsection{The selected question list with \qstep questions}
Table \ref{tab:qstepvsqa} shows the selected question list for \va and \pstep in comparison to step-instructed questions for \qstep. 
\subsubsection{\transpro results with other base models}
Table \ref{tab:model_stab} shows the results of \pstep template using various base models. 

\subsubsection{\transpro results for test set 3: \paran}
Table \ref{tab:res_par3} presents the results for \paran test set. Using GPT-4o-mini as the base model, \transpro consistently outperforms SOTA metrics, showing improvements of up to 0.08 in \kendall and a substantial 17.8-point increase in adequacy. When using the open-source \llama3.3-70b model, \transpro performs marginally below SOTA in correlation but still exceeds the best adequacy SOTA from \cometxxl by more than 10 points.

\subsubsection{Impact of translator votes}\label{apdx:improve_weights}
Table \ref{tab:translator_votes} shows improvements in translator-votes weighted scores compared to non-weighted scores. For \iyyer and \liteval, improvements occur across almost all cases. \paran also shows overall enhancements, except for the \kendall metric.

\begin{figure}
    \centering
    \includegraphics[width=0.95\linewidth]{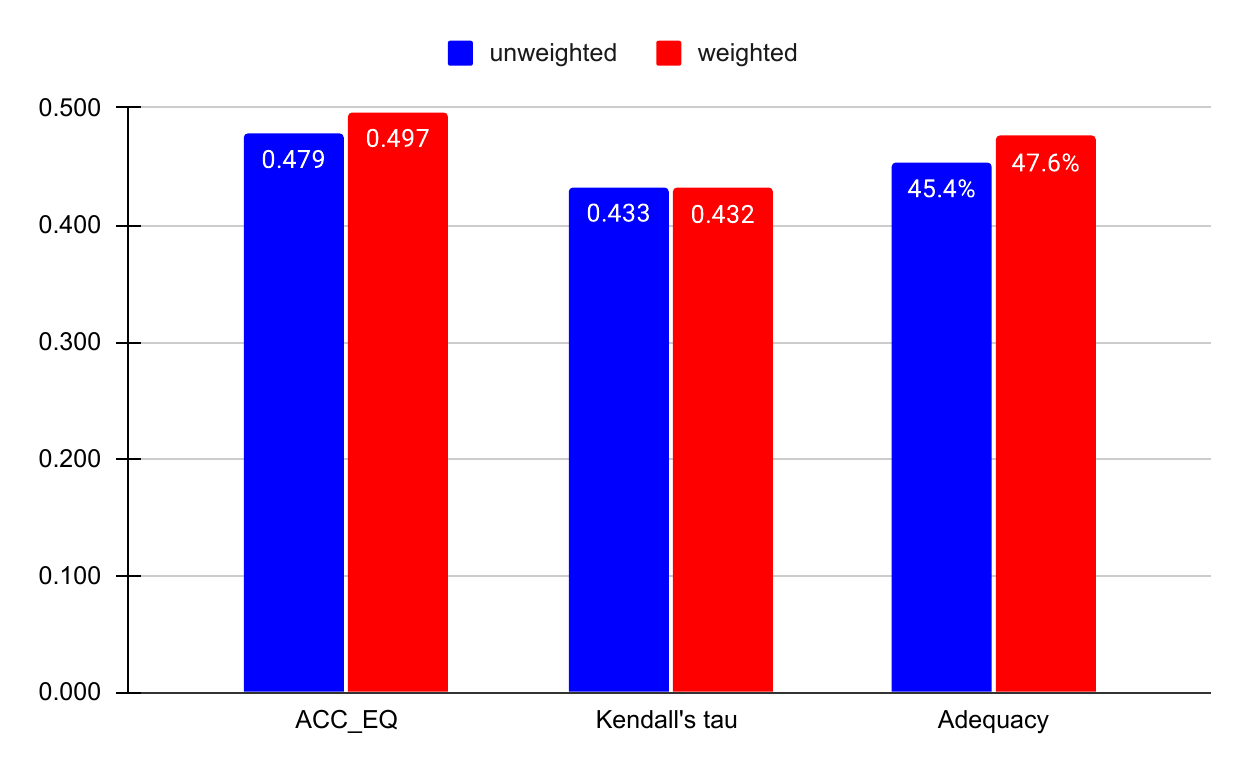}
    \caption{Comparison of weighted vs.\ unweighted \transpro scores: average results across all 3 test datasets.}
    \label{fig:comparison}
\end{figure}

\zr{We evaluate the significance of weighted vs.\ unweighted scores. For the 12 \liteval test cases (4 language pairs × 3 \transpro prompts), 66.7\% of \kendall results and 83.3\% of \acceq results are significant at p < 0.05. For \iyyer (all language pairs combined × 3 prompts), 66.7\% of \kendall results and 100\% of \acceq results reach significance.}
\begin{table}
\tiny
\begin{tabular}{lcccccc}\toprule
\multicolumn{7}{c}{\cellcolor[HTML]{c9daf8}\textbf{Test set 3: \paran}} \\\cmidrule{1-7}
\cellcolor[HTML]{c9daf8}\textbf{Metric} &\multicolumn{2}{c}{\cellcolor[HTML]{c9daf8}\acceq} &\multicolumn{2}{c}{\cellcolor[HTML]{c9daf8}\kendall} &\multicolumn{2}{c}{\cellcolor[HTML]{c9daf8}\makecell{human > MT \\ (\gpttf \& \gtr)}} \\\midrule
\multicolumn{7}{c}{\textbf{SOTA metrics}} \\
\textbf{\gemba} &\multicolumn{2}{c}{0.132} &\multicolumn{2}{c}{-0.014} &\multicolumn{2}{c}{14.7\%} \\
\textbf{\cometkiwi} &\multicolumn{2}{c}{0.231} &\multicolumn{2}{c}{0.155} &\multicolumn{2}{c}{16.7\%} \\
\textbf{\cometxl} &\multicolumn{2}{c}{0.250} &\multicolumn{2}{c}{0.250} &\multicolumn{2}{c}{15.3\%} \\
\textbf{\cometxxl} &\multicolumn{2}{c}{\textbf{0.269}} &\multicolumn{2}{c}{\textbf{0.344}} &\multicolumn{2}{c}{\textbf{29.3\%}} \\
\textbf{M-Prometheus} &\multicolumn{2}{c}{0.119} &\multicolumn{2}{c}{0.148} &\multicolumn{2}{c}{18.0\%} \\
\multicolumn{7}{c}{\textbf{\cometxl Finetuned}} \\
\textbf{\cometxl} &0.250 &$\Delta$ &0.250 &$\Delta$ &15.3\% &$\Delta$ \\
\textbf{\qmono } &0.204 &-0.046 &0.021 &-0.229 &8.0\% &-7.3\% \\
\textbf{\qmulti } &\textbf{0.245} &\textbf{-0.005} &\textbf{0.223} &\textbf{-0.027} &18.7\% &3.3\% \\
\textbf{WMT24} &0.172 &-0.078 &-0.140 &-0.110 &\textbf{22.7\%} &\textbf{7.3\%} \\
\textbf{\qmono +WMT24} &0.179 &-0.071 &0.101 &-0.149 &11.3\% &-4.0\% \\
\textbf{\qmulti + WMT24} &0.224 &-0.026 &0.115 &-0.135 &16.0\% &0.7\% \\
\cellcolor[HTML]{f3f3f3}\textbf{Avg.} &\cellcolor[HTML]{f3f3f3}0.205 &\cellcolor[HTML]{f3f3f3}\textcolor{red}
{-0.045} &\cellcolor[HTML]{f3f3f3}0.064 &\cellcolor[HTML]{f3f3f3}\textcolor{red}
{-0.130} &\cellcolor[HTML]{f3f3f3}14.4\% &\cellcolor[HTML]{f3f3f3}\textcolor{red}
{-0.9\%} \\
\multicolumn{7}{c}{\textbf{\transpro}} \\
\textbf{BEST SOTA} &0.269 &$\Delta$ &0.344 &$\Delta$ &29.3\% &$\Delta$ \\
\multicolumn{7}{c}{\textit{Base model: GPT-4o-mini}} \\
\textbf{\va} &0.266 &-0.003 &0.404 &0.060 &46.7\% &17.3\% \\
\textbf{\vawei} &\textbf{0.268} &\textbf{-0.001} &0.390 &0.047 &\textbf{50.0\%} &\textbf{20.7\%} \\
\textbf{\pstep} &0.265 &-0.005 &\textbf{0.423} &\textbf{0.080} &46.0\% &16.7\% \\
\textbf{\pstepwei} &0.263 &-0.007 &0.403 &0.059 &46.0\% &16.7\% \\
\cellcolor[HTML]{efefef}\textbf{Avg.} &\cellcolor[HTML]{efefef}0.265 &\cellcolor[HTML]{efefef}\textcolor{red}
{-0.004} &\cellcolor[HTML]{efefef}0.405 &\cellcolor[HTML]{efefef}\textcolor{blue}
{0.061} &\cellcolor[HTML]{efefef}47.2\% &\cellcolor[HTML]{efefef}\textcolor{blue}
{17.8\%} \\
\multicolumn{7}{c}{\textit{Base model: LLaMa3.3-70b}} \\
\textbf{\va} &0.219 &-0.050 &0.291 &-0.053 &40.0\% &10.7\% \\
\textbf{\vawei} &0.214 &-0.055 &0.264 &-0.080 &40.0\% &10.7\% \\
\textbf{\pstep} &0.193 &-0.077 &0.270 &-0.074 &39.3\% &10.0\% \\
\textbf{\pstepwei} &0.195 &-0.074 &0.283 &-0.061 &38.7\% &9.3\% \\
\cellcolor[HTML]{efefef}\textbf{Avg.} &\cellcolor[HTML]{efefef}0.205 &\cellcolor[HTML]{efefef}\textcolor{red}
{-0.064} &\cellcolor[HTML]{efefef}0.277 &\cellcolor[HTML]{efefef}\textcolor{red}
{-0.067} &\cellcolor[HTML]{efefef}39.5\% &\cellcolor[HTML]{efefef}\textcolor{blue}
{10.2\%} \\
\bottomrule
\end{tabular}
\caption{Results for test set 3: \paran. \acceq and \kendall measure the segment-level correlation between human judgments (pairwise preference) and metrics. The metric adequacy is reported as the percentage of cases where human translation is scored higher than the outputs from \gpttf and \gtran (\gtr).}\label{tab:res_par3}
\end{table}

\subsection{Qualitative example}
\zr{Table \ref{tab:qualitative} and \ref{tab:qualitative_f} demonstrate qualitative examples from \iyyer with scores from SOTA metrics and \transpro variants. Table \ref{tab:qualitative} shows that SOTA metrics tend to underestimate human translation consistently, while \transpro does not. Table \ref{tab:qualitative_f} demonstrates a failure case of \transpro producing mixed results. The \pstep template successfully ranks human translations higher than MT outputs, while the other two templates (\va and \qstep) fail to make this distinction. Further analysis shows that both failure templates show difficulty with questions about cultural context and cultural translation, often giving these categories lower scores than \pstep. This may occur because \transpro misses subtle stylistic elements when using overly simple \va template, yet struggles with the nuanced understanding when templates become too complex (\qstep). However, human evaluators also report that the differences in translation quality here are only marginal, suggesting that reliably detecting such subtle distinctions requires greater sensitivity that \transpro should further improve upon.}

\clearpage
\thispagestyle{empty}
\begin{figure*}
    \centering
    \begin{subfigure}[t]{0.45\textwidth}
        \centering
        \includegraphics[clip, trim=5cm 0cm 5.8cm -0.cm, height=0.9\textheight]{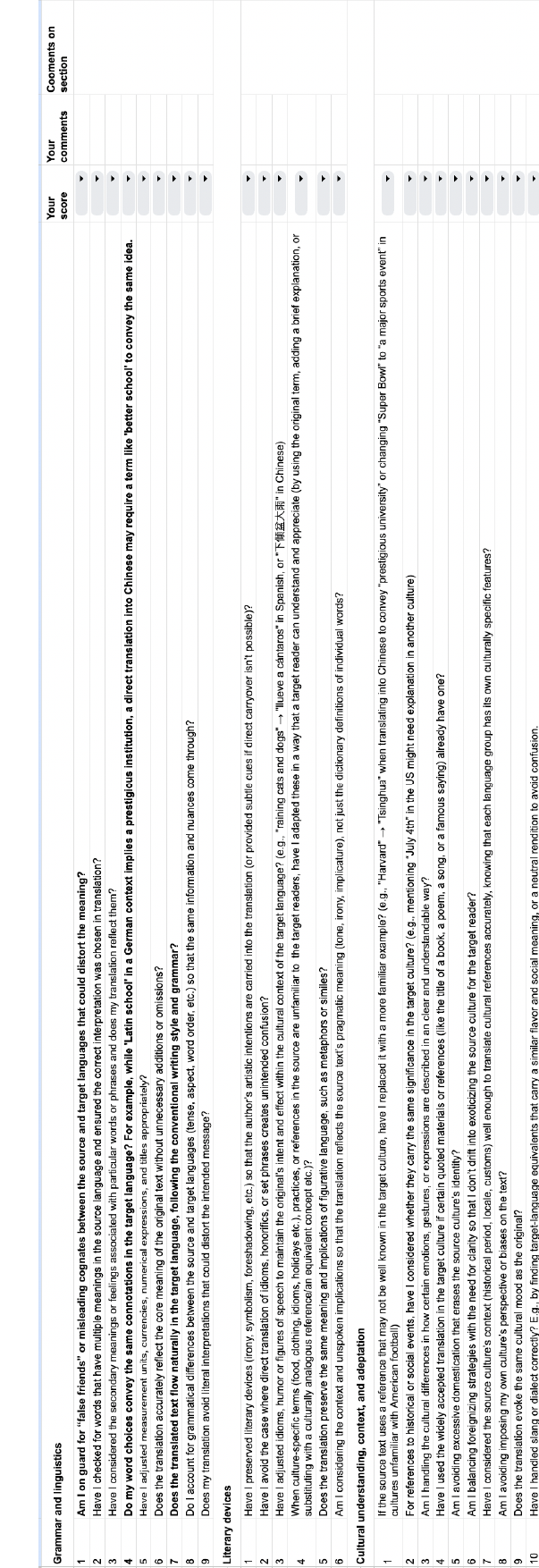}
    \end{subfigure}%
    ~ 
    \begin{subfigure}[t]{0.45\textwidth}
        \centering
        \includegraphics[clip, trim=6cm 0cm 6cm -0.cm, height=0.9\textheight]{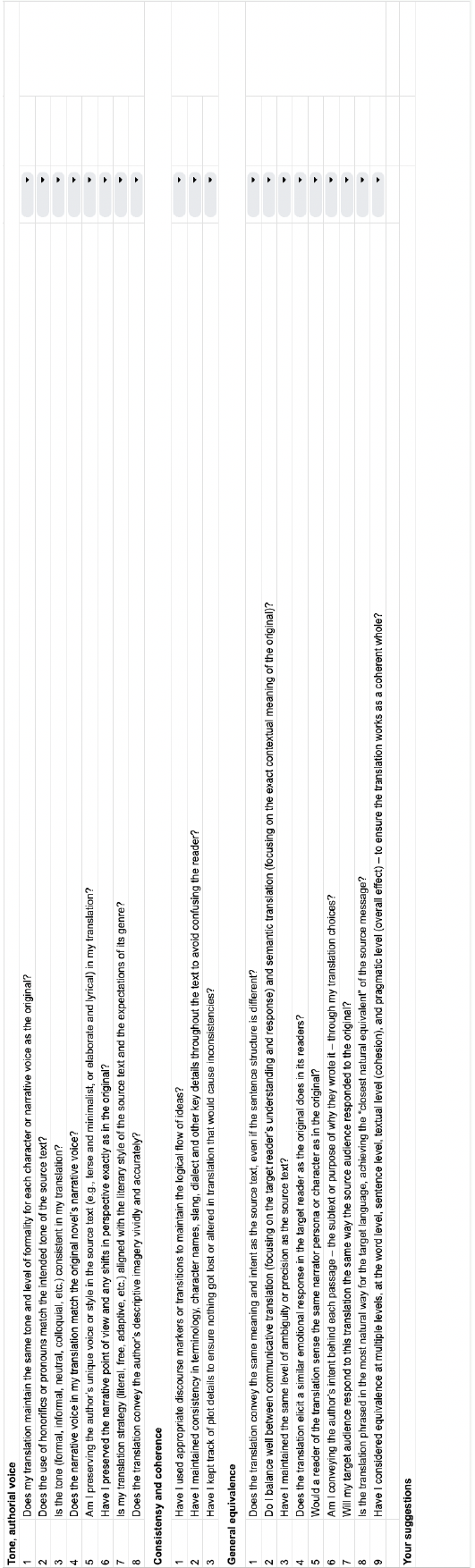}
    \end{subfigure}
    \caption{Screenshot of survey page.}\label{fig:screenqa}
\end{figure*}

\begin{table*}
\scriptsize
\begin{tabular}{m{.15cm}m{.2cm}m{1.1cm}m{9cm}m{.3cm}m{2.2cm}}
\toprule
ID & Asp. &Status &Questions &Score &Note \\
\midrule
35 &CO &S &\cellcolor[HTML]{d9ead3}Have I maintained consistency in terminology, character names, slang, dialect and other key details throughout the text to avoid confusing the reader? &5.00 & \\
2 &GL &R-GI &Have I checked for words that have multiple meanings in the source language and ensured the correct interpretation was chosen in translation? &4.86 &general insensitivity: 91\% yes \\
9 &GL &R-GI &Does my translation avoid literal interpretations that could distort the intended message? &4.86 &general insensitivity: 95\% yes \\
10 &LD &R-GI &Have I preserved literary devices (irony, symbolism, foreshadowing, etc.) so that the author’s artistic intentions are carried into the translation (or provided subtle cues if direct carryover isn’t possible)? &4.86 &general insensitivity: 91\% maybe \\
23 &CCA &R-GI &Am I avoiding imposing my own culture’s perspective or biases on the text? &4.86 &general insensitivity: 97\% yes \\
37 &GE &R-GI &Does the translation convey the same meaning and intent as the source text, even if the sentence structure is different? &4.86 &general insensitivity: 97\% yes \\
1 &GL &S &\cellcolor[HTML]{d9ead3}Am I on guard for ``false friends'' or misleading cognates between the source and target languages that could distort the meaning? &4.86 & \\
29 &TA &S &\cellcolor[HTML]{d9ead3}Does the narrative voice in my translation match the original novel’s narrative voice? &4.86 & \\
7 &GL &S &\cellcolor[HTML]{d9ead3}Does the translated text flow naturally in the target language, following the conventional writing style and grammar? &4.71 & \\
15 &LD &S &\cellcolor[HTML]{d9ead3}Am I considering the context and unspoken implications so that the translation reflects the source text’s pragmatic meaning (tone, irony, implicature), not just the dictionary definitions of individual words? &4.71 & \\
20 &CCA &S &\cellcolor[HTML]{d9ead3}Am I avoiding excessive domestication that erases the source culture’s identity? &4.71 & \\
21 &CCA &S &\cellcolor[HTML]{d9ead3}Am I balancing foreignizing strategies with the need for clarity so that I don’t drift into exoticizing the source culture for the target reader? &4.71 & \\
26 &TA &S &\cellcolor[HTML]{d9ead3}Does my translation maintain the same tone and level of formality for each character or narrative voice as the original? &4.71 & \\
30 &TA &S &\cellcolor[HTML]{d9ead3}Am I preserving the author’s unique voice or style in the source text (e.g., terse and minimalist, or elaborate and lyrical) in my translation? &4.71 & \\
42 &GE &S &\cellcolor[HTML]{d9ead3}Am I conveying the author’s intent behind each passage – the subtext or purpose of why they wrote it – through my translation choices? &4.71 & \\
6 &GL &R-GI &Does the translation accurately reflect the core meaning of the original text without unnecessary additions or omissions? &4.57 &general insensitivity: 96\% yes \\
40 &GE &R-GI &Does the translation elicit a similar emotional response in the target reader as the original does in its readers? &4.57 &general insensitivity: 91\% maybe \\
12 &LD &S &\cellcolor[HTML]{d9ead3}Have I adjusted idioms, humor or figures of speech to maintain the original’s intent and effect within the cultural context of the target language? (e.g., ``raining cats and dogs'' → ``llueve a cántaros'' in Spanish, or \begin{CJK*}{UTF8}{gbsn}``下倾盆大''\end{CJK*} in Chinese) &4.57 & \\
13 &LD &S &\cellcolor[HTML]{d9ead3}When culture-specific terms (food, clothing, idioms, holidays etc.), practices, or references in the source are unfamiliar to the target readers, have I adapted these in a way that a target reader can understand and appreciate (by using the original term, adding a brief explanation, or substituting with a culturally analogous reference/an equivalent concept etc.)? &4.57 & \\
14 &LD &S &\cellcolor[HTML]{d9ead3}Does the translation preserve the same meaning and implications of figurative language, such as metaphors or similes? &4.57 & \\
22 &CCA &S &\cellcolor[HTML]{d9ead3}Have I considered the source culture’s context (historical period, locale, customs) well enough to translate cultural references accurately, knowing that each language group has its own culturally specific features? &4.57 & \\
33 &TA &S &\cellcolor[HTML]{d9ead3}Does the translation convey the author’s descriptive imagery vividly and accurately? &4.57 & \\
36 &CO &S &\cellcolor[HTML]{d9ead3}Have I kept track of plot details to ensure nothing got lost or altered in translation that would cause inconsistencies? &4.43 & \\
3 &GL &R-GI &Have I considered the secondary meanings or feelings associated with particular words or phrases and does my translation reflect them? &4.43 &general insensitivity: 91\% maybe \\
8 &GL &R-GI &Do I account for grammatical differences between the source and target languages (tense, aspect, word order, etc.) so that the same information and nuances come through? &4.43 &general insensitivity: 94\% yes \\
\bottomrule
\end{tabular}
\caption{Question list ranked by translator votes (top 25). ID indicate the original ID in the survey. Asp. stands for the 6 aspects: (1) GL (Grammar \& linguistics); (2) LD (Literary devices); (3) CCA (Cultural understanding, context, \& adaptation); (4) TA (Tone \& authorial voice); (5) CO (Consistency \& coherence); and (6) GE (General equivalence). Status indicates one of four outcomes: (1) S for selected questions, (2) R-GI for rejected questions due to general insensitivity in GPT-4o-mini results (where one answer dominated across translations, showing poor general quality discrimination), (3) R-HI for rejected questions due to human insensitivity (where a high percentage of no/maybe responses for human translations indicated poor discrimination of human translation), and (4) R-TV for questions rejected due to low translator vote scores.}\label{tab:qalist1}
\end{table*}
\begin{table*}
\scriptsize
\begin{tabular}{m{.15cm}m{.2cm}m{1.1cm}m{9cm}m{.3cm}m{2.2cm}}
\toprule
ID & Asp. &Status &Questions &Score &Note \\
\midrule
11 &LD &R-GI &Have I avoid the case where direct translation of idioms, honorifics, or set phrases creates unintended confusion? &4.43 &general insensitivity: 96\% yes \\
4 &GL &S &\cellcolor[HTML]{d9ead3}Do my word choices convey the same connotations in the target language? For example, while 'Latin school' in a German context implies a prestigious institution, a direct translation into Chinese may require a term like 'better school' to convey the same idea. &4.29 & \\
18 &CCA &S &\cellcolor[HTML]{d9ead3}Am I handling the cultural differences in how certain emotions, gestures, or expressions are described in an clear and understandable way? &4.29 & \\
28 &TA &S &\cellcolor[HTML]{d9ead3}Is the tone (formal, informal, neutral, colloquial, etc.) consistent in my translation? &4.29 & \\
31 &TA &S &\cellcolor[HTML]{d9ead3}Have I preserved the narrative point of view and any shifts in perspective exactly as in the original? &4.29 & \\
39 &GE &S &\cellcolor[HTML]{d9ead3}Have I maintained the same level of ambiguity or precision as the source text? &4.29 & \\
41 &GE &S &\cellcolor[HTML]{d9ead3}Would a reader of the translation sense the same narrator persona or character as in the original? &4.29 & \\
38 &GE &R-GI &Do I balance well between communicative translation (focusing on the target reader’s understanding and response) and semantic translation (focusing on the exact contextual meaning of the original)? &4.29 &general insensitivity: 91\% maybe \\
44 &GE &S &\cellcolor[HTML]{d9ead3}Is the translation phrased in the most natural way for the target language, achieving the “closest natural equivalent” of the source message? &4.14 & \\
17 &CCA &R-HI &For references to historical or social events, have I considered whether they carry the same significance in the target culture? (e.g., mentioning “July 4th” in the US might need explanation in another culture) &4.14 &human insensitivity: 100\% no/maybe \\
19 &CCA &R-HI &Have I used the widely accepted translation in the target culture if certain quoted materials or references (like the title of a book, a poem, a song, or a famous saying) already have one? &4.14 &human insensitivity: 22\% no/maybe \\
27 &TA &R-HI &Does the use of honorifics or pronouns match the intended tone of the source text? &4.14 &human insensitivity: 46\% no/maybe \\
24 &CCA &S &\cellcolor[HTML]{d9ead3}Does the translation evoke the same cultural mood as the original? &4.00 & \\
34 &CO &R-GI &Have I used appropriate discourse markers or transitions to maintain the logical flow of ideas? &4.00 &general insensitivity: 94\% yes \\
45 &GE &S &\cellcolor[HTML]{d9ead3}Have I considered equivalence at multiple levels, at the word level, sentence level, textual level (cohesion), and pragmatic level (overall effect) – to ensure the translation works as a coherent whole? &4.00 & \\
32 &TA &R-TV &Is my translation strategy (literal, free, adaptive, etc.) aligned with the literary style of the source text and the expectations of its genre? &3.86 &translator weight < 4 \\
25 &CCA &R-TV &Have I handled slang or dialect correctly? E.g., by finding target-language equivalents that carry a similar flavor and social meaning, or a neutral rendition to avoid confusion. &3.71 &translator weight < 4 \\
5 &GL &R-HI &Have I adjusted measurement units, currencies, numerical expressions, and titles appropriately? &3.43 &human insensitivity: 33\% no/maybe \\
16 &CCA &R-HI &If the source text uses a reference that may not be well known in the target culture, have I replaced it with a more familiar example? (e.g., "Harvard" → "Tsinghua" when translating into Chinese to convey “prestigious university” or changing “Super Bowl” to “a major sports event” in cultures unfamiliar with American football) &3.00 &human insensitivity: 100\% no/maybe \\
43 &GE &R-GI &Will my target audience respond to this translation the same way the source audience responded to the original? &3.00 &general insensitivity: 97\% maybe \\
\bottomrule
\end{tabular}
\caption{Question list ranked by translator votes (26-45). ID indicate the original ID in the survey. Asp. stands for the 6 aspects: (1) GL (Grammar \& linguistics); (2) LD (Literary devices); (3) CCA (Cultural understanding, context, \& adaptation); (4) TA (Tone \& authorial voice); (5) CO (Consistency \& coherence); and (6) GE (General equivalence). Status indicates one of four outcomes: (1) S for selected questions, (2) R-GI for rejected questions due to general insensitivity in GPT-4o-mini results (where one answer dominated across translations, showing poor general quality discrimination), (3) R-HI for rejected questions due to human insensitivity (where a high percentage of no/maybe responses for human translations indicated poor discrimination of human translation), and (4) R-TV for questions rejected due to low translator vote scores.}\label{tab:qalist2}
\end{table*}
\begin{table*}
\centering
\scriptsize
\begin{tabular}{m{.25cm}m{.15cm}m{0.2cm}m{6cm}m{7cm}}
\toprule
Index &ID &Asp. &Questions for \va and \pstep &Step-instructed questions for \qstep \\\midrule
1 &1 &GL &Am I on guard for “false friends” or misleading cognates between the source and target languages that could distort the meaning? &Identify any words that look similar in both languages. Am I on guard for “false friends” or misleading cognates between the source and target languages that could distort the meaning? \\
2 &4 &GL &Do my word choices convey the same connotations in the target language? For example, while 'Latin school' in a German context implies a prestigious institution, a direct translation into Chinese may require a term like 'better school' to convey the same idea. &First, identify all proper names, noun phrases, and cultural/historical references in the source. Then answer: do my word choices convey the same connotations in the target language? Example: While "Latin school" in a German context implies a prestigious institution, a direct translation into Chinese may require a term like "better school" to convey the same idea. \\
3 &7 &GL &Does the translated text flow naturally in the target language, following the conventional writing style and grammar? &Read the translation aloud. Check grammar and syntax. Does the translated text flow naturally in the target language, following conventional writing style and grammar? \\
4 &12 &LD &Have I adjusted idioms, humor or figures of speech to maintain the original’s intent and effect within the cultural context of the target language? (e.g., "raining cats and dogs" → "llueve a cántaros" in Spanish, or \begin{CJK*}{UTF8}{gbsn}``下倾盆大雨''\end{CJK*} in Chinese) & Identify all idioms, jokes, metaphors, unusual expressions, and figures of speech in the source. For each, think about its intended effect or meaning. Have I adjusted them to maintain the original’s intent and effect within the cultural context of the target language? (e.g., "raining cats and dogs" → "llueve a cántaros" in Spanish, or \begin{CJK*}{UTF8}{gbsn}``下倾盆大雨''\end{CJK*} in Chinese) \\
5 &13 &LD &When culture-specific terms (food, clothing, idioms, holidays etc.), practices, or references in the source are unfamiliar to the target readers, have I adapted these in a way that a target reader can understand and appreciate (by using the original term, adding a brief explanation, or substituting with a culturally analogous reference/an equivalent concept etc.)? &First, identify all culture-specific terms (food, clothing, idioms, holidays, etc.), practices, or cultural/historical references in the source. For terms that are unfamiliar to the target readers, answer: have I adapted them in a way that a target reader can understand and appreciate (by using the original term, adding a brief explanation, or substituting with a culturally analogous reference or equivalent concept)? \\
6 &14 &LD &Does the translation preserve the same meaning and implications of figurative language, such as metaphors or similes? &First, identify figurative language such as metaphors, similes, comparisons, or original images in the source. Analyze their meaning and emotional resonance. Does the translation preserve the same meaning and implications? \\
7 &15 &LD &Am I considering the context and unspoken implications so that the translation reflects the source text’s pragmatic meaning (tone, irony, implicature), not just the dictionary definitions of individual words? &Reflect on tone, irony, suggestion, or hidden implications in the source. Does the translation reflect the source text’s pragmatic meaning/function, not just the dictionary definitions of individual words? \\
8 &18 &CCA &Am I handling the cultural differences in how certain emotions, gestures, or expressions are described in an clear and understandable way? &Am I handling cultural differences in how certain emotions, gestures, or expressions are described in a clear and understandable way? \\
9 &20 &CCA &Am I avoiding excessive domestication that erases the source culture’s identity? &Revisit all localized terms. Am I avoiding excessive domestication that erases the source culture’s identity? \\
10 &21 &CCA &Am I balancing foreignizing strategies with the need for clarity so that I don’t drift into exoticizing the source culture for the target reader? &Review translations that feel “foreign” or unusual. Am I balancing foreignizing strategies with the need for clarity so that I don’t drift into exoticizing the source culture for the target reader? \\
11 &22 &CCA &Have I considered the source culture’s context (historical period, locale, customs) well enough to translate cultural references accurately, knowing that each language group has its own culturally specific features? &First, identify all cultural/historical references from the source. Think about the source culture’s context (historical period, locale, customs). Have I considered the context well enough to translate these references accurately, knowing that each language group has its own culturally specific features? \\
12 &24 &CCA &Does the translation evoke the same cultural mood as the original? &First, read the source text alone and summarize the cultural mood. Then read the translation and summarize the cultural mood. Determine: does the translation evoke the same cultural mood as the original? \\
13 &26 &TA &Does my translation maintain the same tone and level of formality for each character or narrative voice as the original? &Does my translation maintain the same tone and level of formality for each character or narrative voice as in the original? \\
14 &28 &TA &Is the tone (formal, informal, neutral, colloquial, etc.) consistent in my translation? &Is the tone (formal, informal, neutral, colloquial, etc.) consistent in my translation? \\
15 &29 &TA &Does the narrative voice in my translation match the original novel’s narrative voice? &Does the narrative voice in my translation match the original novel’s narrative voice? \\
16 &30 &TA &Am I preserving the author’s unique voice or style in the source text (e.g., terse and minimalist, or elaborate and lyrical) in my translation? &Am I preserving the author’s unique voice or style in the source text (e.g., terse and minimalist, or elaborate and lyrical) in my translation? \\
17 &31 &TA &Have I preserved the narrative point of view and any shifts in perspective exactly as in the original? &Note the narrative perspective and any changes (st person, 3rd limited, omniscient). Have I preserved the narrative point of view and any shifts in perspective exactly as in the original? \\
18 &33 &TA &Does the translation convey the author’s descriptive imagery vividly and accurately? &Highlight vivid descriptions and sensory language. Does the translation convey the author’s descriptive imagery vividly and accurately? \\
19 &35 &CO &Have I maintained consistency in terminology, character names, slang, dialect and other key details throughout the text to avoid confusing the reader? &Track key terms, character names, invented words, slang, or dialects. Have I maintained consistency throughout the text to avoid confusing the reader? \\
20 &36 &CO &Have I kept track of plot details to ensure nothing got lost or altered in translation that would cause inconsistencies? &Outline plot developments in the source. Have I kept track of plot details to ensure nothing was lost or altered in translation that would cause inconsistencies? \\
21 &39 &GE &Have I maintained the same level of ambiguity or precision as the source text? &Have I maintained the same level of ambiguity or precision as the source text? \\
22 &41 &GE &Would a reader of the translation sense the same narrator persona or character as in the original? &Understand who the narrator is and their role. Would a reader of the translation sense the same narrator persona or character as in the original? \\
23 &42 &GE &Am I conveying the author’s intent behind each passage – the subtext or purpose of why they wrote it – through my translation choices? &For each passage, ask: why did the author write it this way? Then answer: am I conveying the author’s intent behind each passage—the subtext or purpose of why they wrote it—through my translation choices? \\
24 &44 &GE &Is the translation phrased in the most natural way for the target language, achieving the “closest natural equivalent” of the source message? &Read the translation as a native would. Is it phrased in the most natural way for the target language, achieving the “closest natural equivalent” of the source message? \\
25 &45 &GE &Have I considered equivalence at multiple levels, at the word level, sentence level, textual level (cohesion), and pragmatic level (overall effect) – to ensure the translation works as a coherent whole? &Have I considered equivalence at multiple levels—word level, sentence level, textual level (cohesion), and pragmatic level (overall effect)—to ensure the translation works as a coherent whole? \\
\bottomrule
\end{tabular}
\caption{Selected questions for \va and \pstep vs.\ step-instructed questions for \qstep.}\label{tab:qstepvsqa}
\end{table*}
\begin{table*}\centering
\tiny
\begin{tabular}{m{2.1cm}m{3cm}m{0.75cm}m{.45cm}:m{.4cm}:m{.25cm}:m{.25cm}:m{.25cm}:m{.3cm}:m{.3cm}m{.3cm}:m{.3cm}m{.3cm}:m{.3cm}m{.3cm}}\toprule
\multirow{3}{*}{Source} &\multirow{3}{*}{Target} &\multirow{3}{*}{Model} &\multirow{3}{*}{GEM} &\multirow{3}{*}{KIWI} &\multicolumn{2}{c}{\multirow{2}{*}{\textsc{xcomet}}} &\multirow{3}{*}{M-Pro} &\multirow{3}{*}{TRE} &\multicolumn{6}{c}{\transpro} \\\cmidrule{10-15}
& & & & & & & & &\multicolumn{2}{c}{\va} &\multicolumn{2}{c}{\pstep}\textbf{}\textbf{} &\multicolumn{2}{c}{\qstep} \\\cmidrule{6-7}\cmidrule{10-15}
& & & & &XL &XXL & & &\textbf{-} &\textbf{w} &\textbf{-} &\textbf{w} &\textbf{-} &\textbf{w} \\\midrule
\multirow{4}{*}{\makecell[{{m{2.1cm}}}]{Bis auf Selmas Schwägerin Elsbeth waren die Leute im Dorf meistens nicht abergläubisch. Sie machten unbekümmert all das, was man bei Aberglauben nicht machen darf: Sie saßen gelassen unter Wanduhren, obwohl man bei Aberglauben daran sterben kann, sie schliefen mit dem Kopf zur Tür hin, obwohl das bei Aberglauben bedeutet, dass man durch genau die Tür bald mit den Füßen zuerst hinausgetragen wird. Sie hängten zwischen Weihnachten und Neujahr Wäsche auf, was, wie Elsbeth warnte, bei Aberglauben einem Suizid oder einer Beihilfe zum Mord gleichkommt. Sie erschraken nicht, wenn nachts das Käuzchen rief, wenn ein Pferd im Stall stark schwitzte, wenn ein Hund nachts jaulte, mit gesenktem Kopf.}} &Except for Selma’s sister-in-law Elsbeth, people in the village were for the most part not superstitious. They blithely broke all of superstition’s rules: They sat calmly under wall clocks even though the superstitious can die from it. They slept with their heads toward the door, though superstition claimed they’d soon be carried out that very door feetfirst. They hung laundry to dry between Christmas and New Year’s, which, according to superstition, Elsbeth would remind them, amounts to suicide or accessory to murder. They were not frightened when owls hooted, when a horse in the stall broke into a heavy sweat, when a dog howled in the night with its head lowered. &Human &\textcolor{red}{-7.0} &\textcolor{red}{0.71} &\textcolor{red}{0.67} &\textcolor{red}{0.54} &\textbf{\textcolor{blue}{6}} &54.1 &\textbf{\textcolor{blue}{4.4}} &\textbf{\textcolor{blue}{3.99}} &\textbf{\textcolor{blue}{4.7}} &\textbf{\textcolor{blue}{4.24}} &\textbf{\textcolor{blue}{4.8}} &\textbf{\textcolor{blue}{4.33}} \\
\cmidrule{2-15}
&Apart from Selma's sister-in-law Elsbeth, most people in the village were not superstitious. They did all the things that superstition forbids: they sat calmly under wall clocks, even though superstition says that it can kill you; they slept with their heads towards the door, even though superstition says that it means you will be carried out of the door feet first; they hung out laundry between Christmas and New Year, which, as Elsbeth warned, was akin to suicide or aiding and abetting murder according to superstition. They were not scared when the screech owl called at night, when a horse sweated heavily in the stable, or when a dog howled with its head down at night. &\makecell{\gpttf \\ \para} &\textbf{\textcolor{blue}{-6.0}} &0.75 &0.79 &\textbf{\textcolor{blue}{0.66}} &\textbf{\textcolor{blue}{6}} &55.2 &\textbf{\textcolor{blue}{4.4}} &3.97 &4.4 &3.96 &4.1 &3.71 \\
\cmidrule{2-15}
&Except for Selma's sister-in-law Elsbeth, the people in the village were mostly not superstitious. They did all the things that one should not do in superstition: They sat calmly under wall clocks, even though in superstition one can die from that, they slept with their head towards the door, even though in superstition that means that one will soon be carried out of the door feet first. They hung out laundry between Christmas and New Year's, which, as Elsbeth warned, in superstition is equivalent to suicide or aiding in murder. They did not shudder when the owl called at night, when a horse was sweating heavily in the stable, or when a dog howled with its head down. &\makecell{\gpttf \\ \sentp} &\textcolor{red}{-7.0} &\textbf{\textcolor{blue}{0.77}} &\textbf{\textcolor{blue}{0.82}} &\textbf{\textcolor{blue}{0.66}} &\textbf{\textcolor{blue}{6}} &\textbf{\textcolor{blue}{59.4}} &\textcolor{red}{3.7} &\textcolor{red}{3.36} &\textcolor{red}{4.0} &3.63 &\textcolor{red}{3.8} &\textcolor{red}{3.44} \\
\cmidrule{2-15}
&Apart from Selma's sister-in-law Elsbeth, most people in the village were not superstitious. They carelessly did all the things that superstition forbids: They sat calmly beneath wall clocks, even though superstition says that it can kill you, and they slept with their heads towards the door, even though superstition says that you will soon be carried out of it feet first. They hung laundry between Christmas and New Year's, which, as Elsbeth warned, was considered superstitious to be equivalent to suicide or aiding in murder. They were not startled when the owl called out at night, when a horse sweated heavily in the stable, or when a dog howled with its head bowed at night. &\makecell{\gpttf \\ \sent}&\textbf{\textcolor{blue}{-6.0}} &0.75 &0.79 &0.60 &4 &\textcolor{red}{51.7} &4.3 &3.89 &\textcolor{red}{4.0} &\textcolor{red}{3.62} &3.9 &3.52 \\
\bottomrule
\end{tabular}
\caption{Qualitative example from \iyyer. We use abbreviated forms of metrics: GEM (\gemba), KIWI (\cometkiwi), XL (\cometxl), XXL (\cometxxl), M-Pro (M-Prometheus), and TRE (TREQA-QE). W indicates the weighted version of \transpro. We highlight the highest score of individual metrics in \textbf{\textcolor{blue}{blue}} and the lowest in \textcolor{red}{red}.}\label{tab:qualitative}
\end{table*}
\begin{table*}\centering
\tiny
\begin{tabular}{m{2.1cm}m{3cm}m{0.75cm}m{.45cm}:m{.4cm}:m{.25cm}:m{.25cm}:m{.25cm}:m{.3cm}:m{.3cm}m{.3cm}:m{.3cm}m{.3cm}:m{.3cm}m{.3cm}}\toprule
\multirow{3}{*}{Source} &\multirow{3}{*}{Target} &\multirow{3}{*}{Model} &\multirow{3}{*}{GEM} &\multirow{3}{*}{KIWI} &\multicolumn{2}{c}{\multirow{2}{*}{\textsc{xcomet}}} &\multirow{3}{*}{M-Pro} &\multirow{3}{*}{TRE} &\multicolumn{6}{c}{\transpro} \\\cmidrule{10-15}
& & & & & & & & &\multicolumn{2}{c}{\va} &\multicolumn{2}{c}{\pstep}\textbf{}\textbf{} &\multicolumn{2}{c}{\qstep} \\\cmidrule{6-7}\cmidrule{10-15}
& & & & &XL &XXL & & &\textbf{-} &\textbf{w} &\textbf{-} &\textbf{w} &\textbf{-} &\textbf{w} \\\midrule
\multirow{4}{*}{\makecell[{{m{2.1cm}}}]{Das stimmte. Der Hund war schneematschfarben, er war verwaschen grau und zottelig wie ein ausschließlicher Irischer Wolfhund ohne etwas anderes drin. Sein Körper war noch klein, aber seine Pfoten waren groß wie Bärentatzen, und wir wussten alle, was das bedeutete. Selma stand immer noch erhoben vor der Küchenbank. Sie schaute lange auf den Hund. Dann sah sie meinen Vater an, als sei er ein Geschenkideengeschäft.}} &It was true. The dog was the color of slush. It was watery gray and shaggy as only purebred Irish wolfhounds can be. Its body was still small, but its paws were as big as a bear’s, and we all knew what that meant. Selma was still standing in front of the kitchen bench. She looked at the dog for a long time. Then she looked at my father as if he were a gift shop. &human &\textbf{\textcolor{blue}{-7}} &\textcolor{red}{0.69} &\textbf{\textcolor{blue}{0.76}} & \textbf{\textcolor{blue}{0.79}} &\textbf{\textcolor{blue}{4}} &\textcolor{red}{53.9} &\textcolor{red}{3.2} &3.26 &\textbf{\textcolor{blue}{4.1}} &\textbf{\textcolor{blue}{3.71}} &3.3 &2.98 \\
\cmidrule{2-15}
&That was true. The dog was snow-muddy-colored, he was washed-out grey and shaggy like an exclusive Irish Wolfhound without anything else mixed in. His body was still small, but his paws were as big as bear's claws, and we all knew what that meant. Selma was still standing tall in front of the kitchen bench. She looked at the dog for a long time. Then she looked at my father as if he was a gift shop. &\makecell{\gpttf \\ \para} &\textcolor{red}{-37} &0.76 &\textcolor{red}{0.75} &0.71 &\textcolor{red}{2} &65.3 &3.4 &\textcolor{red}{3.08} &\textcolor{red}{3.6} &\textcolor{red}{3.27} &\textbf{\textcolor{blue}{3.5}} &\textbf{\textcolor{blue}{3.17}} \\
\cmidrule{2-15}
&That was true. The dog was snow-mushroom colored, he was washed-out grey and shaggy like a pure Irish Wolfhound with nothing else mixed in. His body was still small, but his paws were big like bear paws, and we all knew what that meant. Selma still stood erect in front of the kitchen bench. She looked at the dog for a long time. Then she looked at my father as if he was a gift shop. &\makecell{\gpttf \\ \sentp} &\textcolor{red}{-37} &\textbf{\textcolor{blue}{0.77}} &\textcolor{red}{0.75} &\textbf{\textcolor{blue}{0.79}} &\textbf{\textcolor{blue}{4}} &\textbf{\textcolor{blue}{67.9}} &3.4 &\textcolor{red}{3.08} &\textcolor{red}{3.6} &\textcolor{red}{3.27} &\textcolor{red}{3.2} &\textcolor{red}{2.88} \\
\cmidrule{2-15}
&That's true. The dog was snow-muddy in color, he was washed-out gray and shaggy like an exclusive Irish wolfhound with nothing else mixed in. His body was still small, but his paws were as big as a bear's, and we all knew what that meant. Selma was still standing in front of the kitchen counter. She looked at the dog for a long time. Then she looked at my father as if he were a gift shop. &\makecell{\gpttf \\ \sent} &\textbf{\textcolor{blue}{-7}} &0.76 &\textcolor{red}{0.75} &\textcolor{red}{0.69} &\textcolor{red}{2} &60.5 &\textbf{\textcolor{blue}{3.6}} &\textbf{\textcolor{blue}{3.27}} &\textcolor{red}{3.6} &\textcolor{red}{3.27} &3.4 &3.07 \\
\bottomrule
\end{tabular}
\caption{Qualitative failure example from \iyyer. We use abbreviated forms of metrics: GEM (\gemba), KIWI (\cometkiwi), XL (\cometxl), XXL (\cometxxl), M-Pro (M-Prometheus), and TRE (TREQA-QE). W indicates the weighted version of \transpro. We highlight the highest score of individual metrics in \textbf{\textcolor{blue}{blue}} and the lowest in \textcolor{red}{red}.}\label{tab:qualitative_f}
\end{table*}
\end{document}